\documentclass[10pt,twocolumn,letterpaper]{article}

\usepackage{cvpr}
\usepackage{times}
\usepackage{epsfig}
\usepackage{graphicx}
\usepackage{amsmath}
\usepackage{amssymb}
\usepackage{placeins}
\usepackage{tabularx}
\usepackage{subcaption}
\usepackage{helvet}

\newcommand{\fig}[1]{Figure~\ref{fig:#1}}

\newcommand{\tab}[1]{Table~\ref{tab:#1}}

\newcommand{\eq}[1]{(\ref{eq:#1})}

\usepackage[pagebackref=true,breaklinks=true,letterpaper=true,colorlinks,bookmarks=false]{hyperref}

\cvprfinalcopy 

\def\cvprPaperID{5428} 

\begin{document}

\title{Textured Neural Avatars}

\author{Aliaksandra Shysheya $^{1,2}$ \quad Egor Zakharov $^{1,2}$ \quad Kara-Ali Aliev $^1$ \quad Renat Bashirov $^1$ \and Egor Burkov $^{1,2}$ \quad Karim Iskakov $^1$ \quad Aleksei Ivakhnenko $^1$ \quad Yury Malkov $^1$ \and Igor Pasechnik $^1$ \quad Dmitry Ulyanov $^{1,2}$ \quad Alexander Vakhitov $^{1,2}$ \quad  Victor Lempitsky $^{1,2}$
\vspace{1.5mm}\\
$^1 $Samsung AI Center, Moscow \quad $^2 $Skolkovo Institute of Science and Technology, Moscow
}

\thispagestyle{empty}

\twocolumn[{%
\renewcommand\twocolumn[1][]{#1}%
\maketitle
\vspace*{-2em}
\begin{center}
    \centering
    \includegraphics[width=\textwidth]{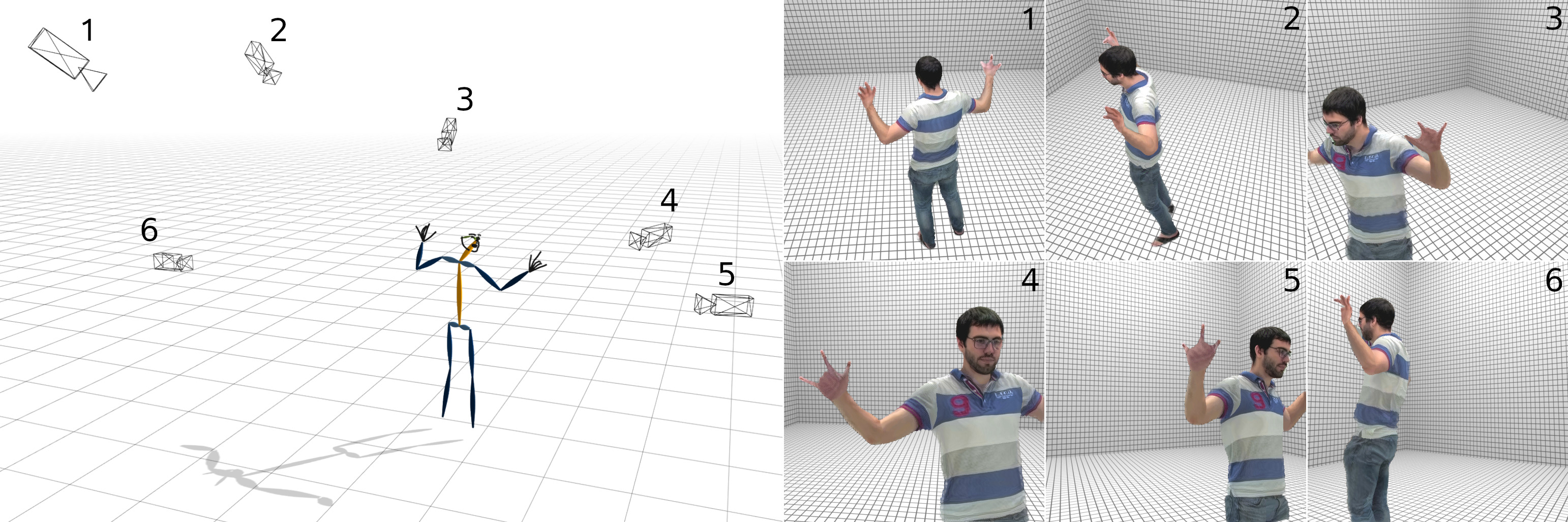}\\
    \includegraphics[width=\textwidth]{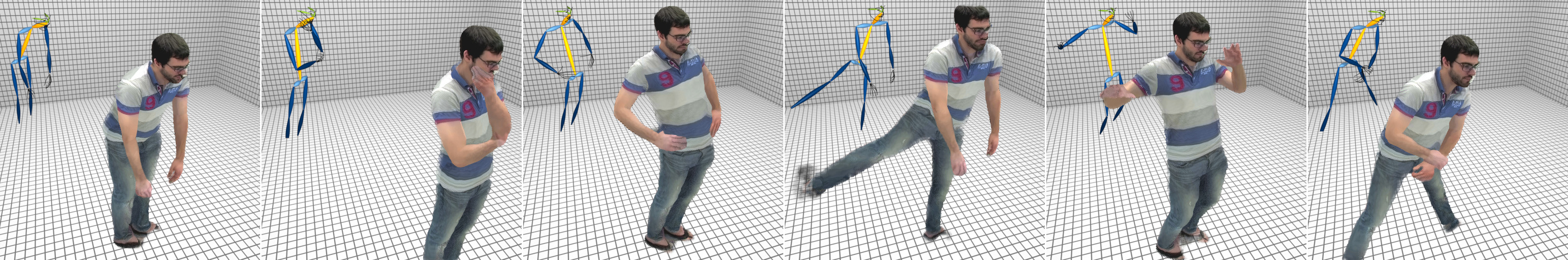}
    \captionof{figure}{We propose a new model for neural rendering of humans. The model is trained for a single person and can produce renderings of this person from novel viewpoints (top) or in the new body pose (bottom) unseen during training. To improve generalization, our model retains explicit texture representation, which is learned alongside the rendering neural network.}
    \label{fig:teaser}
\end{center}
}]

\begin{abstract}
We present a system for learning full-body \emph{neural avatars}, i.e.\ deep networks that produce full-body renderings of a person for varying body pose and camera position. Our system takes the middle path between the classical graphics pipeline and the recent deep learning approaches that generate images of humans using image-to-image translation. In particular, our system estimates an explicit two-dimensional texture map of the model surface. At the same time, it abstains from explicit shape modeling in 3D. Instead, at test time, the system uses a fully-convolutional network to directly map the configuration of body feature points w.r.t.\ the camera  to the 2D texture coordinates of individual pixels in the image frame. We show that such a system  is capable of learning to generate realistic renderings while being trained on videos annotated with 3D poses and foreground masks. We also demonstrate that maintaining an explicit texture representation helps our system to achieve better generalization compared to systems that use direct image-to-image translation. 
\end{abstract}

\section{Introduction}


Capturing and rendering human body in all of its complexity under varying pose and imaging conditions is one of the core problems of both computer vision and computer graphics. Recently, there is a surge of interest that involves deep convolutional networks (ConvNets) as an alternative to traditional computer graphics means. Realistic \textit{neural rendering} of body fragments e.g.\ faces~\cite{Kim18,Lombardi18,Suwajanakorn17}, eyes~\cite{Ganin16}, hands~\cite{Mueller18} is now possible. Very recent works have shown the abilities of such networks to generate views of a person with a varying body pose but with a fixed camera position, and using an excessive amount of training data~\cite{Aberman18,Chan18,Liu18,Wang18}. In this work, we focus on the learning of \textit{neural avatars}, i.e.\ generative deep networks that are capable of rendering views of individual people under varying body pose defined by a set of 3D positions of the body joints and under varying camera positions (\fig{teaser}). We prefer to use body joint positions to represent the human pose, as joint positions are often easier to capture using marker-based or marker-less motion capture systems. 

Generally, neural avatars can serve as an alternative to classical (``neural-free'') avatars based on a standard computer graphics pipeline that estimates a user-personalized body mesh in a neutral position, performs skinning (deformation of the neutral pose), and projects the resulting 3D surface onto the image coordinates, while superimposing person-specific 2D texture. Neural avatars attempt to shortcut the multiple stages of the classical pipeline and to replace them with a single network that learns the mapping from the input (the location of body joints) to the output (the 2D image).  As a part of our contribution, we demonstrate that, however appealing for its conceptual simplicity, existing pose-to-image translation networks generalize poorly to new camera views, and therefore new architectures for neural avatars are required.



Towards this end, we present a neural avatar system that does full-body rendering and combines the ideas from the classical computer graphics, namely the decoupling of geometry and texture, with the use of deep convolutional neural networks. In particular, similarly to the classic pipeline, our system explicitly estimates the 2D textures of body parts. The 2D texture within the classical pipeline effectively transfers the appearance of the body fragments across camera transformations and body articulations. Keeping this component within the neural pipeline boosts generalization across such transforms. The role of the convolutional network in our approach is then confined to predicting the texture coordinates of individual pixels in the output 2D image given the body pose and the camera parameters (\fig{textured}). Additionally, the network predicts the body foreground/background mask. 

In our experiments, we compare the performance of our \textit{textured neural avatar} with a direct video-to-video translation approach~\cite{Wang18}, and show that explicit estimation of textures brings additional generalization capability and improves the realism of the generated images for new views and/or when the amount of training data is limited. 
\section{Related work}
\label{sect:related}

Our approach is closely related to a vast number of previous works, and below we discuss a small subset of these connections.

Building \textbf{full-body avatars} from image data has long been one of the main topics of computer vision research. Traditionally, an avatar is defined by a 3D geometric mesh of a certain neutral pose, a texture, and a skinning mechanism that transforms the mesh vertices according to pose changes. A large group of works has been devoted to body modeling from 3D scanners \cite{PonsMoll15}, registered multi-view sequences \cite{Robertini17} as well as from depth and RGB-D sequences \cite{Bogo15,Weiss11,Yu18}. On the other extreme are methods that fit skinned parametric body models to single images \cite{Balan08,Bogo16,Hasler10, Kanazawa18,Omran18,Pavlakos18,Starck03}. Finally, research on building full-body avatars from monocular videos has started~\cite{Alldieck18b,Alldieck18a}. Similarly to the last group of works, our work builds an avatar from a video or a set of unregistered monocular videos. The classical (computer graphics) approach to modeling human avatars requires explicit physically-plausible modeling of human skin, hair, sclera, clothing surface, as well as 
motion under pose changes. Despite considerable progress in reflectivity modeling \cite{alexander2010digital,Donner08,Klehm15,Weyrich06,Wood15} and better skinning/dynamic surface modeling \cite{Feng15,Loper15,Stavness14}, the computer graphics approach still requires considerable ``manual'' effort of designers
to achieve high realism~\cite{alexander2010digital} and to pass the so-called uncanny valley~\cite{Mori70}, especially if real-time rendering of avatars is required. 

\textbf{Image synthesis using deep convolutional neural networks} is a thriving area of research \cite{Dosovitskiy15,Goodfellow14} and a lot of recent effort has been directed onto synthesis of realistic human faces~\cite{Choi18,Karras18,Sungatullina18}. Compared to traditional computer graphics representations, deep ConvNets model data by fitting an excessive number of learnable weights to training data. Such ConvNets avoid explicit modeling of the surface geometry, surface reflectivity, or surface motion under pose changes, and therefore do not suffer from the lack of realism of the corresponding components. On the flipside, the lack of ingrained geometric or photometric models in this approach means that generalizing to new poses and in particular to new camera views may be problematic.
Still a lot of progress has been made over the last several years for the neural modeling of personalized talking head models \cite{Kim18,Lombardi18,Suwajanakorn17}, hair \cite{Wei18}, hands \cite{Mueller18}. Notably, the recent system~\cite{Lombardi18} has achieved very impressive results for neural face rendering, while decomposing view-dependent texture and 3D shape modeling.  

Over the last several months, several groups have presented results of neural modeling of full bodies \cite{Aberman18,Chan18,Liu18,Wang18}. While the presented results are very impressive, the approaches still require a large amount of training data. They also assume that the test images are rendered with the same camera views as the training data, which in our experience makes the task considerably simpler than modeling body appearance from an arbitrary viewpoint. In this work, we aim to expand the neural body modeling approach to tackle the latter, harder task. The work~\cite{Martin18} uses a combination of classical and neural rendering to render human body from new viewpoints, but does so based on depth scans and therefore with a rather different algorithmic approach.

A number of recent works \textbf{warp a photo of a person} to a new photorealistic image with modified gaze direction \cite{Ganin16}, modified facial expression/pose \cite{Cao18,Shu18,Tulyakov18,Wiles18}, or modified body pose \cite{Balakrishnan18,Neverova18,Siarohin18,Tulyakov18}, whereas the warping field is estimated using a deep convolutional network (while the original photo effectively serves as a texture). These approaches are however limited in their realism and/or the amount of change they can model, due to their reliance on a single photo of a given person for its input. Our approach also disentangles texture from surface geometry/motion modeling but trains from videos, therefore being able to handle harder problem (full body multi-view setting) and to achieve higher realism.

Our system relies on the \textbf{DensePose} body surface parameterization (UV parameterization) similar to the one used in the classical graphics-based representation. Part of our system performs a mapping from the body pose to the surface parameters (UV coordinates) of image pixels. This makes our approach related to the DensePose approach \cite{Guler18} and the earlier works \cite{Guler17,Taylor12} that predict UV coordinates of image pixels from the input photograph. Furthermore, our approach uses DensePose results~\cite{Guler18} for pretraining.

Our system is  related to approaches that extract \textbf{textures from multi-view image collections}~\cite{Goldlucke09,Lempitsky07} or multi-view video collections \cite{Volino14} or a single video \cite{RavAcha08}. Our approach is also related to free-viewpoint video compression and rendering systems, e.g.~\cite{Casas14,Collet15,Dou17,Volino14}. Unlike those works, ours is restricted to scenes containing a single human. At the same time, our approach aims to generalize not only to new camera views but also to new user poses unseen in the training videos. The work of \cite{Xu11} is the most related to ours in this group, as they warp the individual frames of the multi-view video dataset according to the target pose to generate new sequences. The poses that they can handle, however, are limited by the need to have a close match in the training set, which is a strong limitation given the combinatorial nature of the human pose configuration space.

\section{Methods}

\newcommand{\floor}[1]{\lfloor #1 \rfloor}
\newcommand{\ceil}[1]{\lceil #1 \rceil}
\renewcommand{\l}{\mathcal{L}}
\newcommand{\Strut}{\rule[-.4\baselineskip]{0pt}{\baselineskip}}

\paragraph{Notation.} We use the lower index $i$ to denote objects that are specific to the $i$-th training or test image. We use uppercase notation, e.g.\ $B_i$ to denote a stack of maps (a third-order tensor/three-dimensional array) corresponding to the $i$-th training or test image. We use the upper index to denote a specific map (channel) in the stack, e.g.\ $B^j_i$. Furthermore, we use square brackets to denote elements corresponding to a specific image location, e.g.\ $B^j_i[x,y]$ denotes the scalar element in the $j$-th map of the stack $B_i$ located at location $(x,y)$, and $B_i[x,y]$ denotes the vector of elements corresponding to all maps sampled at location $(x,y)$.

\paragraph{Input and output.} In general, we are interested in synthesizing images of a certain person given her/his pose. We assume that the pose for the $i$-th image comes in the form of 3D joint positions defined in the camera coordinate frame. As an input to the network, we then consider a map stack $B_i$, where each map $B_i^j$ contains the rasterized $j$-th segment (bone) of the ``stickman'' (skeleton) projected on the camera plane. To retain the information about the third coordinate of the joints, we linearly interpolate the depth value between the joints defining the segments, and use the interpolated values to define the values in the map $B_i^j$ corresponding to the bone pixels (the pixels not covered by the $j$-th bone are set to zero). Overall, the stack $B_i$ incorporates the information about the person and the camera pose.

As an output of the whole system, we expect an RGB image (a three-channel stack) $I_i$ and a single channel mask $M_i$, defining the pixels that are covered by the avatar. Below, we consider two approaches: the \textit{direct translation} baseline, which directly maps $B_i$ into $\{I_i,M_i\}$ and the \textit{textured neural avatar} approach that performs such mapping indirectly using texture mapping.

In both cases, at training time, we assume that for each input frame $i$, the input joint locations and the ``ground truth'' foreground mask are estimated, and we use 3D body pose estimation and human semantic segmentation to extract them from raw video frames. At test time, given a real or synthetic background image $\tilde I_i$, we generate the final view by first predicting $M_i$ and $I_i$ from the body pose and then linearly blending the resulting avatar into an image: $\hat I_i= I_i \odot M_i + \tilde I_i \odot (1-M_i)$ (where $\odot$ defines a ``location-wise'' product, i.e. the RGB values at each location are multiplied by the mask value at this location). 

\paragraph{Direct translation baseline.}  The direct approach that we consider as a baseline to ours is to learn an image translation network that maps the map stack $B^k_i$ to the map stacks $I_i$ and $M_i$ (usually the two output stacks are produced within two branches that share the initial stage of the processing \cite{Dosovitskiy15}). Generally, mappings between stacks of maps can be implemented using fully-convolutional architectures. Exact architectures and losses for such networks is an active area of research \cite{Chen17,Isola17,Johnson16,Ulyanov16}. Very recent works \cite{Aberman18,Chan18,Liu18,Wang18} have used direct translation (with various modifications) to synthesize the view of a person for a fixed camera. We use the video-to-video variant of this approach~\cite{Wang18} as a baseline for our method. 


\begin{figure*}
    \vspace{-1cm}
    \centering
    \includegraphics[width=\textwidth]{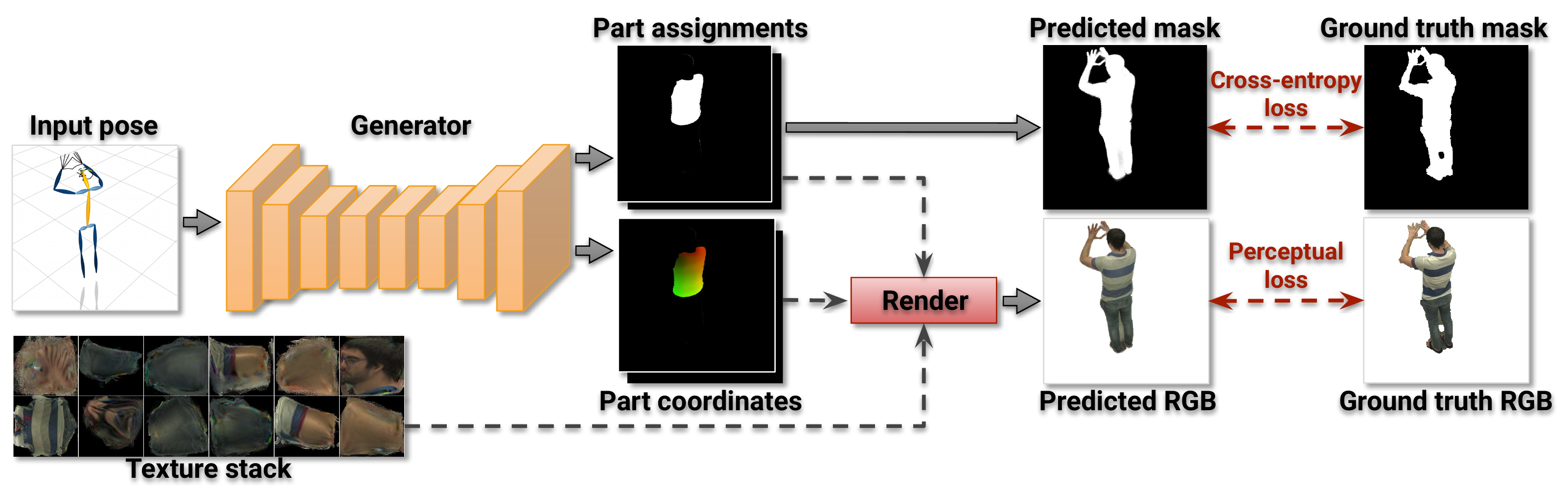}
    \caption{The overview of the textured neural avatar system. The input pose is defined as a stack of "bone" rasterizations (one bone per channel; here we show it as a skeleton image). The input is processed by the fully-convolutional network (generator) to produce the body part assignment map stack and the body part coordinate map stack. These stacks are then used to sample the body texture maps at the locations prescribed by the part coordinate stack with the weights prescribed by the part assignment stack to produce the RGB image. In addition, the last body assignment stack map corresponds to the background probability. During learning, the mask and the RGB image are compared with ground-truth and the resulting losses are backpropagated through the sampling operation into the fully-convolutional network and onto the texture, resulting in their updates.}
    \label{fig:textured}
\end{figure*}

\paragraph{Textured neural avatar.} The direct translation approach relies on the generalization ability of ConvNets and incorporates very little domain-specific knowledge into the system. As an alternative, we suggest the textured avatar approach, that explicitly estimates the textures of body parts, thus ensuring the similarity of the body surface appearance under varying pose and cameras. 

Following the DensePose approach \cite{Guler18}, we subdivide the body into $n{=}24$ parts, where each part has a 2D parameterization. Each body part also has the texture map $T^k$, which is a color image of a fixed pre-defined size (256${\times}$256 in our implementation). The training process for the textured neural avatar estimates personalized part parameterizations and textures.

Again, following the DensePose approach, we assume that each pixel in an image of a person is (soft)-assigned to one of $n$ parts or to the background and with a specific location on the texture of that part (body part coordinates). Unlike DensePose, where part assignments and body part coordinates are induced from the image,  our approach at test time aims to predict them based solely on the pose $B_i$. 

The introduction of the body surface parameterization outlined above changes the translation problem. For a given pose defined by $B_i$, the translation network now has to predict the stack $P_i$ of body part assignments and the stack $C_i$ of body part coordinates, where $P_i$ contains $n{+}1$ maps of non-negative numbers that sum to identity (i.e. $\sum_{k=0}^{n} P^k_i[x,y] = 1$ for any position $(x, y)$), and $C_i$ contains $2n$ maps of real numbers between 0 and $w$, where $w$ is the spatial size (width and height) of the texture maps $T^k$. 

The map channel $P_i^k$ for $k=0,\dots,n{-}1$ is then interpreted as the probability of the pixel to belong to the $k$-th body part, and the map channel $P_i^n$ corresponds to the probability of the background. The coordinate maps $C_i^{2k}$ and $C_i^{2k+1}$ correspond to the pixel coordinates on the $k$-th body part. Specifically, once the part assignments $P_i$ and body part coordinates $C_i$ are predicted, the image $I_i$ at each pixel $(x,y)$ is reconstructed as a weighted combination of texture elements, where the weights and texture coordinates are prescribed by the part assignment maps and the coordinate maps correspondingly:
\begin{align}
s(P_i,C_i,T)[x,y] &= \sum_{k=0}^{n-1} P_i^k[x,y]\cdot\nonumber\\ &T^k\left[C_i^{2k}[x,y],C_i^{2k+1}[x,y]\right]\,,\label{eq:texturing}
\end{align}
where $s(\cdot,\cdot,\cdot)$ is the sampling function (layer) that outputs the RGB map stack given the three input arguments.
In \eq{texturing}, the texture maps $T^k$ are sampled at non-integer locations $(C_i^{2k}[x,y],C_i^{2k+1}[x,y])$ in a piecewise-differentiable manner using bilinear interpolation \cite{Jaderberg15}.


When training the neural textured avatar, we learn a convolutional network $g_\phi$ with learnable parameters $\phi$ to translate the input map stacks $B_i$ into the body part assignments and the body part coordinates. As $g_\phi$ has two branches (``heads''), we denote with $g_\phi^P$ the branch that produces the body part assignments stack, and with $g_\phi^C$ the branch that produces the body part coordinates. To learn the parameters of the textured neural avatar, we optimize the loss between the generated image and the ground truth image $\bar I_i$:
\begin{equation}
    \l_\text{image}(\phi,T) = d_\text{Image}\left(\Strut\bar I_i,\; s\left(g_\phi^P(B_i),g_\phi^C(B_i),T\right)\right)\,\label{eq:imageloss}
\end{equation}
where $d_\text{Image}(\cdot,\cdot)$ is a loss used to compare two images. In our current implementation we use a simple perceptual loss \cite{Gatys16,Johnson16,Ulyanov16}, which computes the maps of activations within pretrained fixed VGG network~\cite{Simonyan14} for both images and evaluates the L1-norm between the resulting maps (\texttt{Conv1,6,11,20,29} of VGG19 were used). More advanced adversarial losses~\cite{Goodfellow14} popular in image translation \cite{Dosovitskiy16,Isola17} can also be used here.

During the stochastic optimization, the gradient of the loss \eq{imageloss} is backpropagated through \eq{texturing} both into the translation network $g_\phi$ and onto the texture maps $T^k$, so that minimizing this loss updates not only the network parameters but also the textures themselves. As an addition, the learning also optimizes the mask loss that measures the discrepancy between the ground truth background mask $1-\bar M_i$ and the background mask prediction:
\begin{equation}
    \l_\text{mask}(\phi,T) = d_\text{BCE}\left(\Strut\bar 1-M_i,\; g_\phi^P(B_i)^n \right)\,\label{eq:maskloss}
\end{equation}
where $d_\text{BCE}$ is the binary cross-entropy loss, and $g_\phi^P(B_i)^n$ corresponds to the $n$-th (i.e.\ background) channel of the predicted part assignment map stack. After backpropagation of the weighted combination of \eq{imageloss} and \eq{maskloss}, the network parameters $\phi$ and the textures maps $T^k$ are updated. As the training progresses, the texture maps change (\fig{textured}), and so does the body part coordinate predictions, so that the learning is free to choose the appropriate parameterization of body part surfaces.

\paragraph{Initialization of textured neural avatar.} The success of our network depends on the initialization strategy. When training from multiple video sequences, we use the DensePose system~\cite{Guler18} to initialize the textured neural avatar. Specifically, we run DensePose on the training data and pretrain $g_\phi$ as a translation network between the pose stacks $B_i$ and the DensePose outputs. 

An alternative way that is particularly attractive when training data is scarce is to initialize the avatar is through transfer learning. In this case, we simply take $g_\phi$ from another avatar trained on abundant data. The explicit decoupling of geometry from appearance in our method facilitates transfer learning, as the geometrical mapping provided by the network $g_\phi$ usually does not need to change much between two people, especially if the body types are not too dissimilar. 

\begin{figure}
    \centering
    \includegraphics[width=0.45\textwidth]{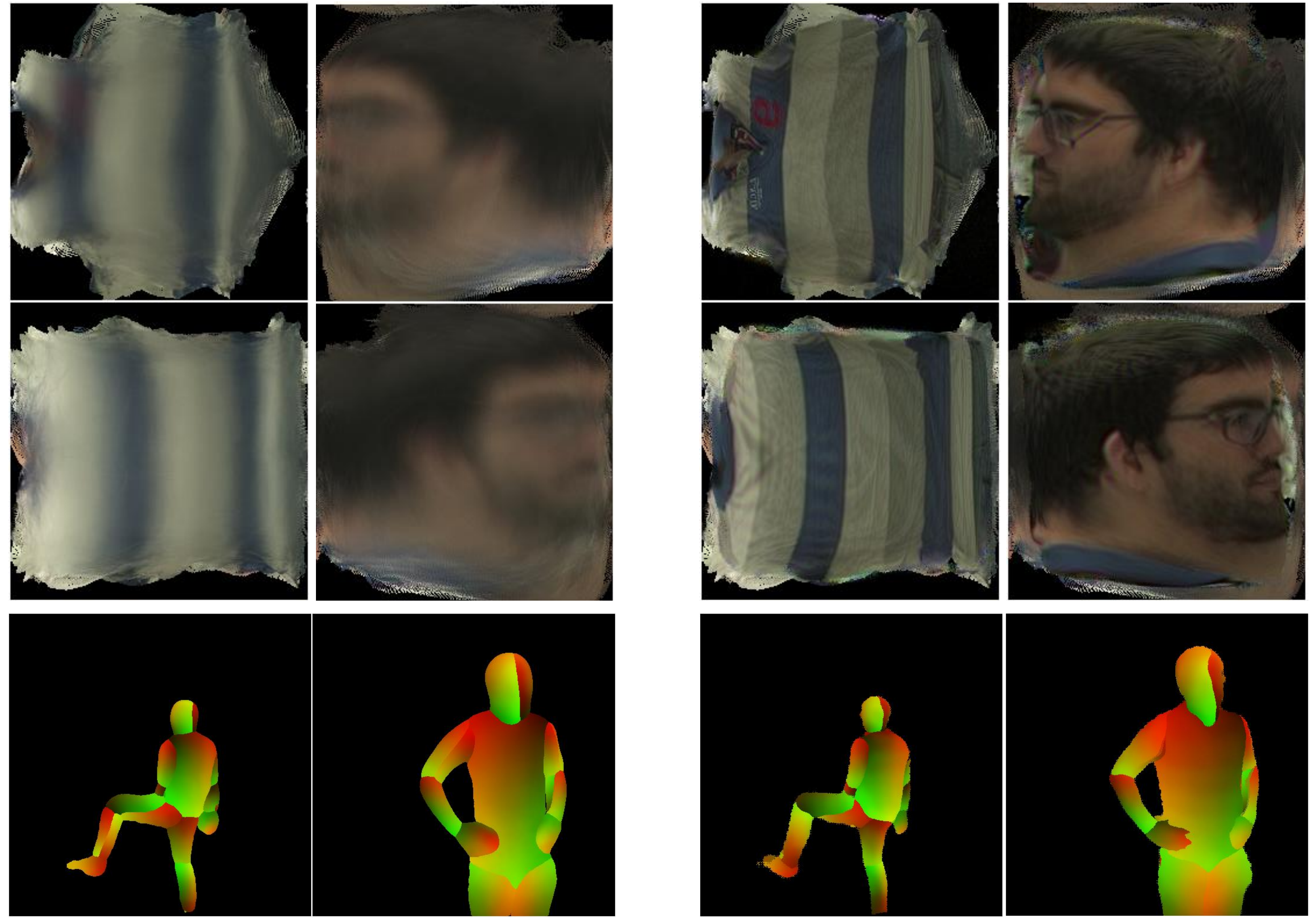}
    \caption{The impact of the learning on the texture (top, shown for the same subset of maps $T^k$) and on the convolutional network $g_\phi^C$ predictions (bottom, shown for the same pair of input poses). Left part shows the starting state (after initialization), while the right part shows the final state, which is considerably different from the start.}
    \label{fig:progress}
\end{figure}

Once the mapping $g_\phi$ has been initialized, the texture maps $T^k$ are initialized as follows. Each pixel in the training image is assigned to a single body part (according to the prediction of the pretrained $g_\phi^P$) and to a particular texture pixel on the texture of the corresponding part (according to the prediction of the pretrained $g_\phi^C$). Then, the value of each texture pixel is initialized to the mean of all image pixels assigned to it (the texture pixels assigned zero pixels are initialized to black). The initialized texture $T$ and $g_\phi$ usually produce images that are only coarsely reminding the person, and they change significantly during the end-to-end learning (\fig{progress}).



\section{Experiments}
\begin{figure*}
\includegraphics[width=\textwidth]{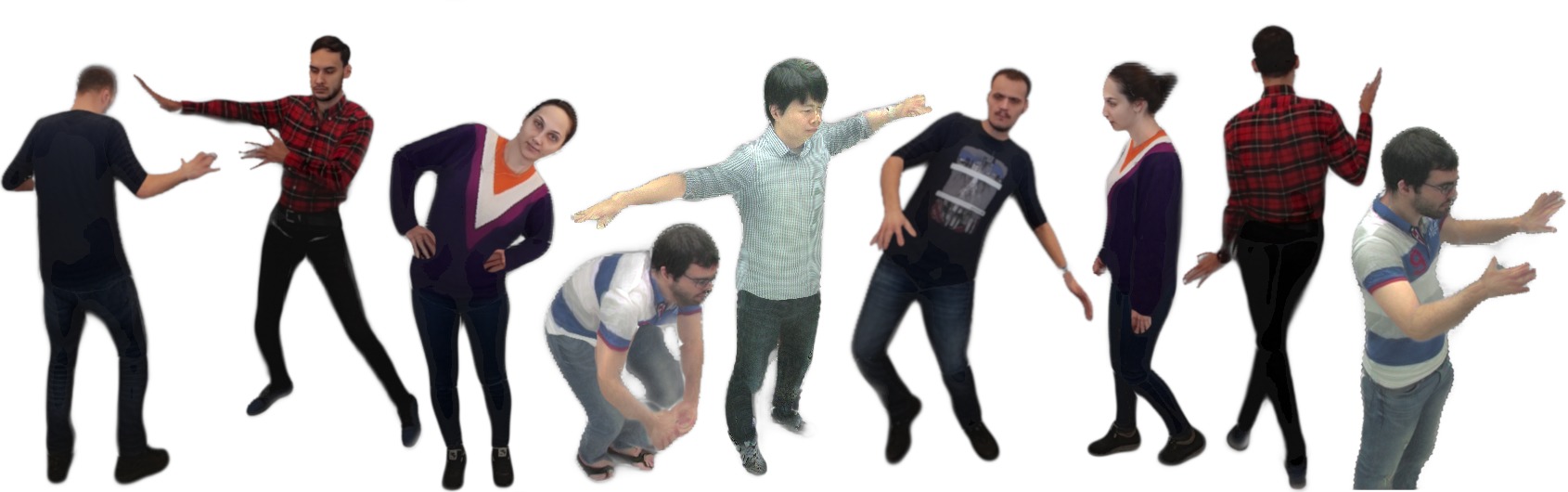}
    \caption{Renderings produced by multiple textured neural avatars (for all people in our study). All renderings are produced from the new viewpoints unseen during training.}
    \label{fig:assorti}
\end{figure*}

\begin{table*}
    \begin{subtable}{\textwidth}
        \centering
        \begin{tabular}{
                c | 
                *{2}{>{\centering\arraybackslash}p{2cm}} | 
                *{3}{>{\centering\arraybackslash}p{1.15cm}} | 
                *{3}{>{\centering\arraybackslash}p{1.15cm}}
            }
            & \multicolumn{2}{c|}{(a) User study} 
            & \multicolumn{3}{c|}{(b) SSIM score}
            & \multicolumn{3}{c}{(c) Frechet distance} \\
            \cline{2-9}
                                 & Ours-v-V2V & Ours-v-Direct   & V2V & Direct & Ours  & V2V & Direct & Ours   \\ 
            \hline
            CMU1-16  & 0.56    & 0.75  & $0.908$ & $0.899$      & 0.919   & 6.7     & 7.3         & 8.8 \\
            CMU2-16  & 0.54    & 0.74     & $0.916$ & $0.907$      & 0.922   & 7.0     & 8.8         & 10.7 \\
            CMU1-6   & 0.50    & 0.92  & $0.905$ & $0.896$      & 0.914   & 7.7     & 10.7         & 8.9\\
            CMU2-6   & 0.53    & 0.71     & $0.918$ & $0.907$      & 0.920   & 7.0     & 9.7         & 10.4 \\
        \end{tabular}
    \end{subtable}
    \caption{Quantitative comparison of the three models operating on different datasets (see text for discussion).}\label{tab:tables}
\end{table*}

Below, we discuss the details of the experimental validation, provide comparison with baseline approaches, and show qualitative results. The project webpage\footnote{\url{https://saic-violet.github.io/texturedavatar/}} contains more videos of the learned avatars.


\paragraph{Architecture.}

We input 3D pose via bone rasterizations, where each bone, hand and face are drawn in separate channels. We then use standard image translation architecture~\cite{Johnson16} to perform a mapping from these bones' rasterizations to texture assignments and coordinates. This architecture consists of downsampling layers, stack of residual blocks, operating at low dimensional feature representations, and upsampling layers. We then split the network into two roughly equal parts: encoder and decoder, with texture assignments and coordinates having separate decoders. We use 4 downsampling and upsampling layers with initial 32 channels in the convolutions and 256 channels in the residual blocks. The ConvNet $g_\phi$ has 17 million parameters. 


\paragraph{Datasets.} We train neural avatars on several types of datasets. First, we consider collections of multi-view videos registered in time and space, where 3D pose estimates can be obtained via triangulation of 2D poses. We use two subsets (corresponding to two persons from the 171026\textunderscore pose2 scene) from the CMU Panoptic dataset collection ~\cite{Joo_2017_TPAMI}, referring to them as \texttt{CMU1} and \texttt{CMU2} (both subsets have approximately four minutes~/~7,200 frames in each camera view). We consider two regimes: training on 16 cameras (\texttt{CMU1-16} and \texttt{CMU2-16}) or six cameras (\texttt{CMU1-6} and \texttt{CMU2-6}). The evaluation is done on the hold-out cameras and hold-out parts of the sequence (no overlap between train and test in terms of the cameras or body motion).

We have also captured our own multi-view sequences of three subjects using a rig of seven cameras, spanning approximately $30^\circ$. In one scenario, the training sets included six out of seven cameras, where the duration of each video was approximately six minutes (11,000 frames). We show qualitative results for the hold-out camera as well as from new viewpoints. In the other scenario described below, training was done based on a video from a single camera.

\begin{figure*}[h!]
    \centering
    \includegraphics[width=0.89\textwidth]{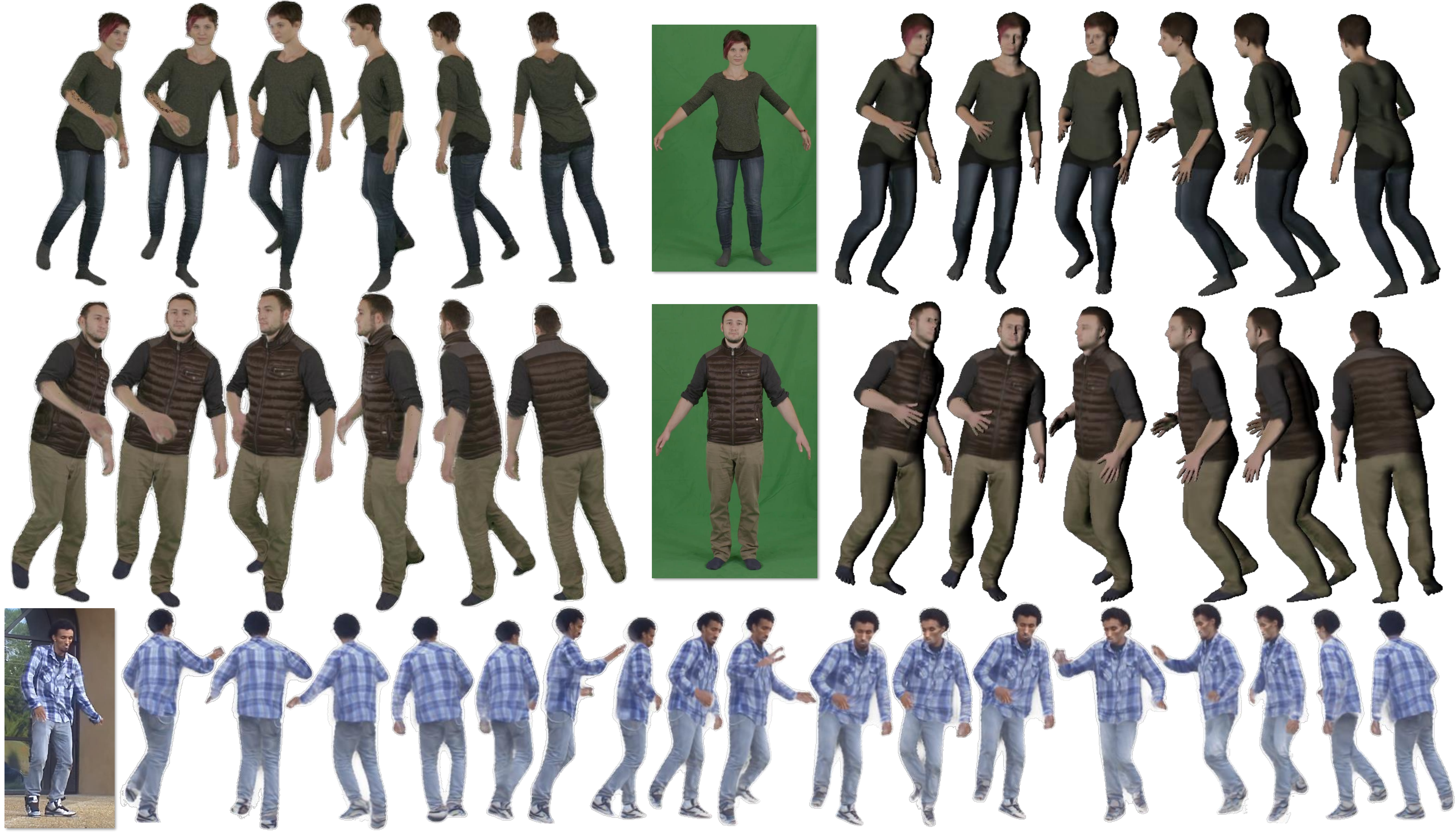}
    \caption{Results on external monocular sequences. Rows 1-2: avatars for sequences from \cite{Alldieck18a}  in an unseen pose (left -- ours, right -- \cite{Alldieck18a}). Row 3 -- the textured avatar computed from a popular YouTube video ('PUMPED UP KICKS DUBSTEP'). In general, our system is capable of learning avatars from monocular videos.}\label{fig:extra}
\end{figure*}

Finally, we evaluate on two short monocular sequences from \cite{Alldieck18a} and a Youtube video in \fig{extra}.

\paragraph{Pre-processing.} Our system expects 3D human pose as input. For non-CMU datasets, we used the OpenPose-compatible~\cite{cao2017realtime, simon2017hand} 3D pose formats, represented by 25~body joints, 21~joints for each hand and 70~facial landmarks. For the CMU Panoptic datasets, we use the available 3D pose annotation as input (which has 19 rather than 25 body joints). To get a 3D pose for non-CMU sequences we first apply the OpenPose 2D pose estimation engine to five consecutive frames of the monocular RGB image sequence. Then we concatenate and lift the estimated 2D poses to infer the 3D pose of the last frame by using a multi-layer perceptron model. The perceptron is trained on the CMU 3D pose annotations (augmented with position of the feet joints by triangulating the output of OpenPose) in orthogonal projection. 

For foreground segmentation we use DeepLabv3+ with Xception-65 backbone~\cite{deeplabv3plus2018} initially trained on PASCAL VOC 2012~\cite{Everingham15} and fine-tuned on HumanParsing dataset~\cite{ATR,CO-CNN} to predict initial human body segmentation masks. We additionally employ GrabCut~\cite{Rother04} with background/foreground model initialized by the masks to refine object boundaries on the high-resolution images. Pixels covered by the skeleton rasterization were always added to the foreground mask.

\begin{figure*}[h!]
    \vspace{-0.5cm}
    \captionsetup[subfigure]{labelformat=empty}
    
    \centering
      \begin{subfigure}{0.25\columnwidth}
      \includegraphics[trim={0cm 0cm 6cm 4cm},clip,width=\textwidth]{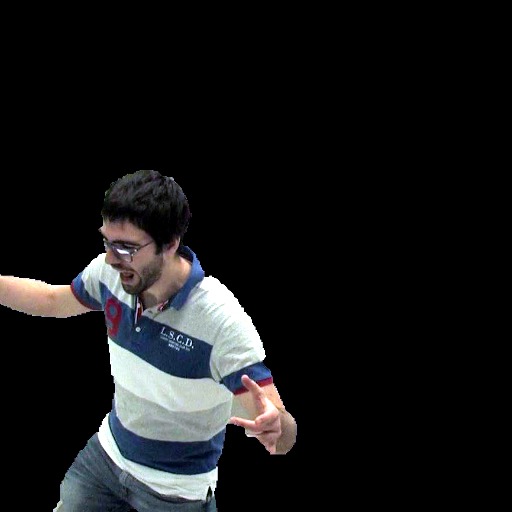}  
      \end{subfigure}
      \begin{subfigure}{0.25\columnwidth}
      \includegraphics[trim={0cm 0cm 5cm 2.5cm},clip,width=\textwidth]{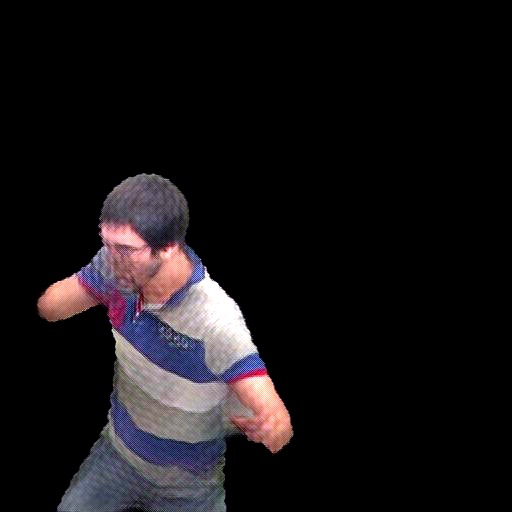} 
      \end{subfigure}
      \begin{subfigure}{0.25\columnwidth}
      \includegraphics[trim={0cm 0cm 5cm 2.5cm},clip,width=\textwidth]{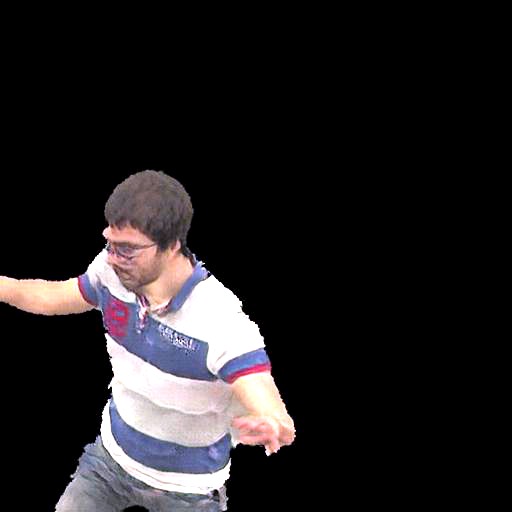} 
      \end{subfigure}
      \begin{subfigure}{0.25\columnwidth}
      \includegraphics[trim={0cm 0cm 5cm 2.5cm},clip,width=\textwidth]{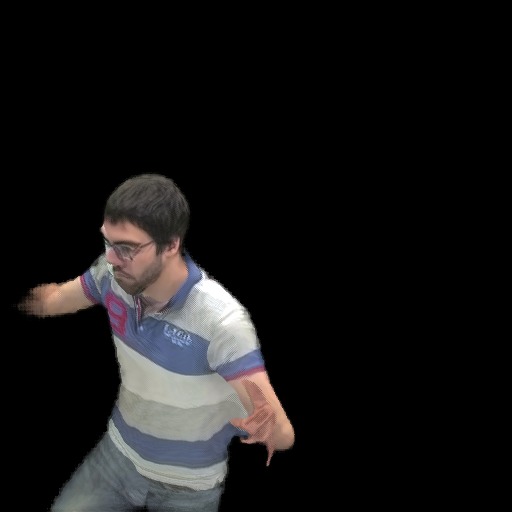}
      \end{subfigure}\hspace*{0.5em}
      \begin{subfigure}{0.25\columnwidth}
      \includegraphics[trim={0cm 0cm 5cm 3cm},clip,width=\textwidth]{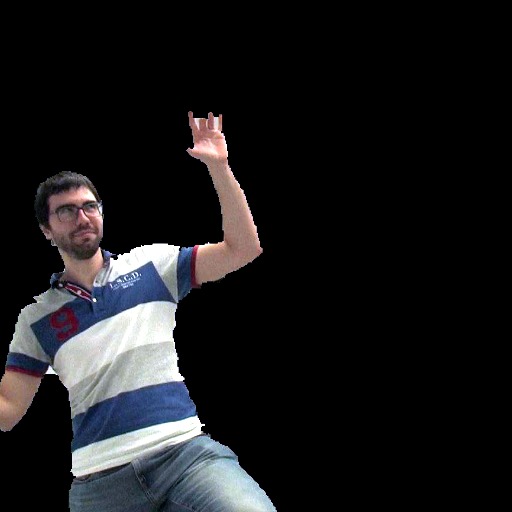}  
      \end{subfigure}
      \begin{subfigure}{0.25\columnwidth}
      \includegraphics[trim={0cm 0cm 5cm 2.5cm},clip,width=\textwidth]{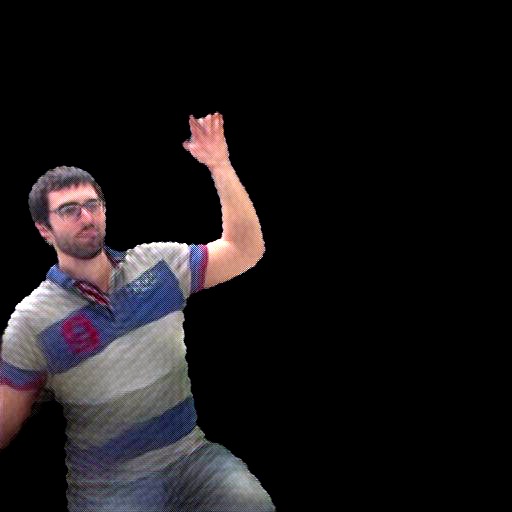} 
      \end{subfigure}
      \begin{subfigure}{0.25\columnwidth}
      \includegraphics[trim={0cm 0cm 5cm 2.5cm},clip,width=\textwidth]{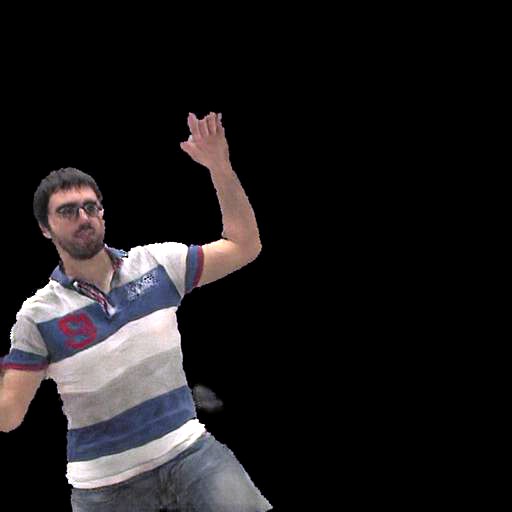}
      \end{subfigure}
      \begin{subfigure}{0.25\columnwidth}
      \includegraphics[trim={0cm 0cm 5cm 2.5cm},clip,width=\textwidth]{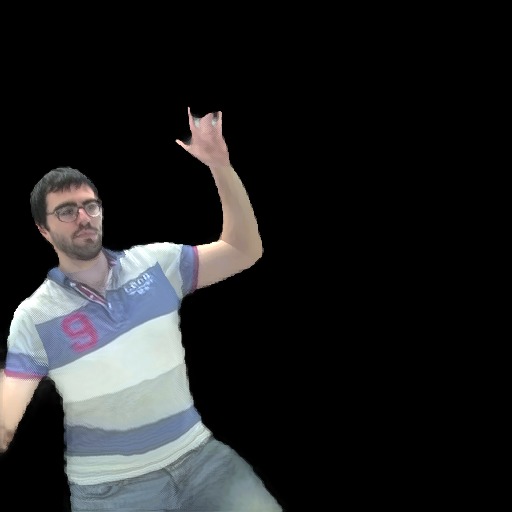}
      \end{subfigure}   \\
      \begin{subfigure}{0.25\columnwidth}
      \includegraphics[trim={3cm 0cm 4cm 5cm},clip,width=\textwidth]{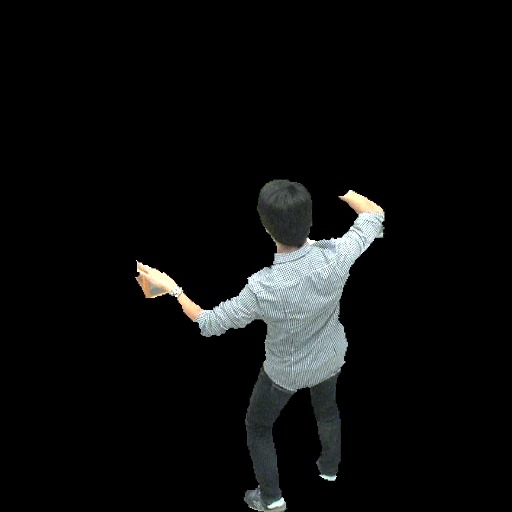} 
      \caption{GT}
      \end{subfigure}
      \begin{subfigure}{0.25\columnwidth}
      \includegraphics[trim={3cm 0cm 4cm 5cm},clip,width=\textwidth]{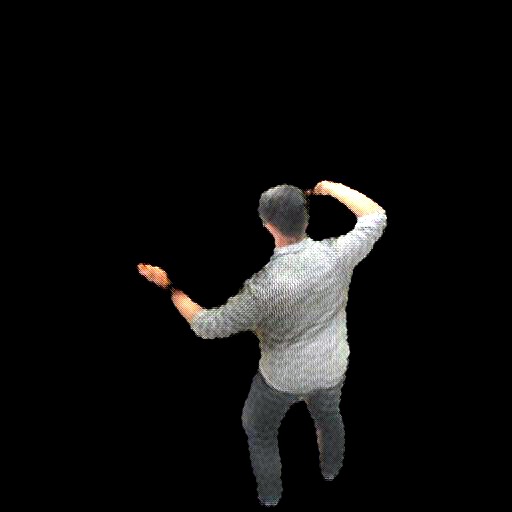} 
      \caption{Direct}
      \end{subfigure}
      \begin{subfigure}{0.25\columnwidth}
      \includegraphics[trim={3cm 0cm 4cm 5cm},clip,width=\textwidth]{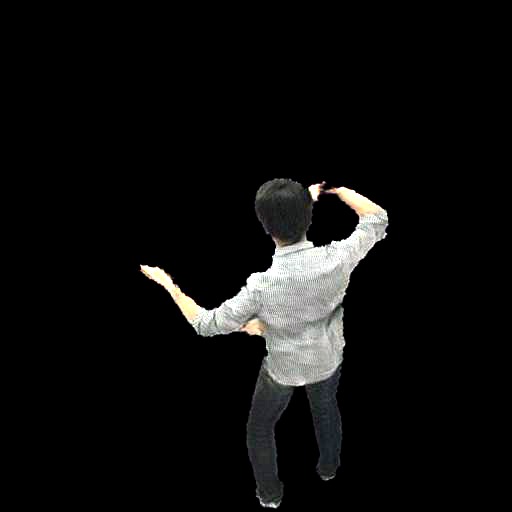} 
      \caption{V2V}
      \end{subfigure}
      \begin{subfigure}{0.25\columnwidth}
      \includegraphics[trim={3cm 0cm 4cm 5cm},clip,width=\textwidth]{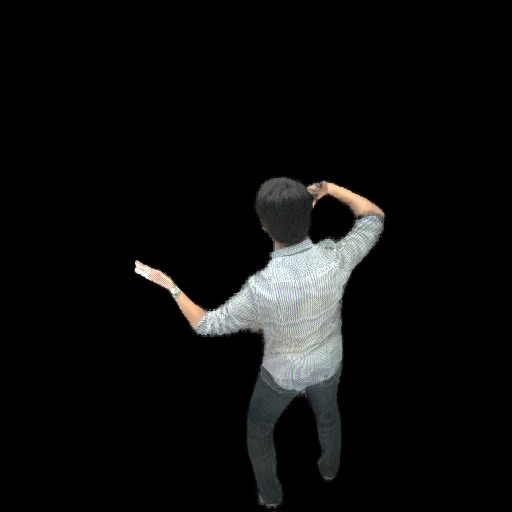}
      \caption{Proposed}
      \end{subfigure}\hspace*{0.5em}      
      \begin{subfigure}{0.25\columnwidth}
      \includegraphics[trim={0cm 0cm 5cm 2.5cm},clip,width=\textwidth]{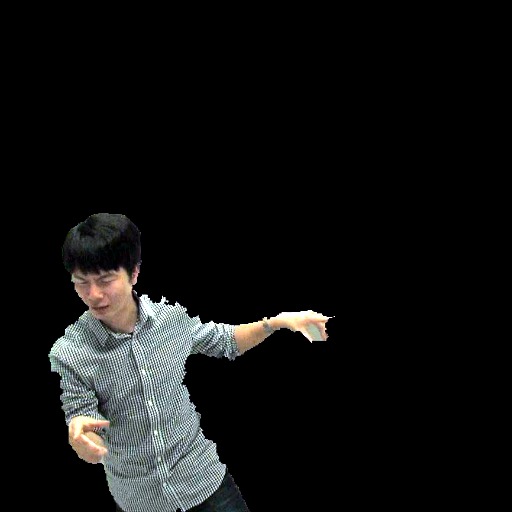} 
      \caption{GT}
      \end{subfigure}
      \begin{subfigure}{0.25\columnwidth}
      \includegraphics[trim={0cm 0cm 5cm 2.5cm},clip,width=\textwidth]{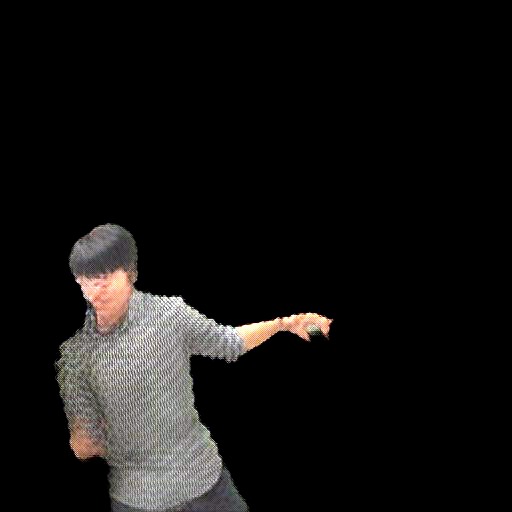}
      \caption{Direct}
      \end{subfigure}
      \begin{subfigure}{0.25\columnwidth}
      \includegraphics[trim={0cm 0cm 5cm 2.5cm},clip,width=\textwidth]{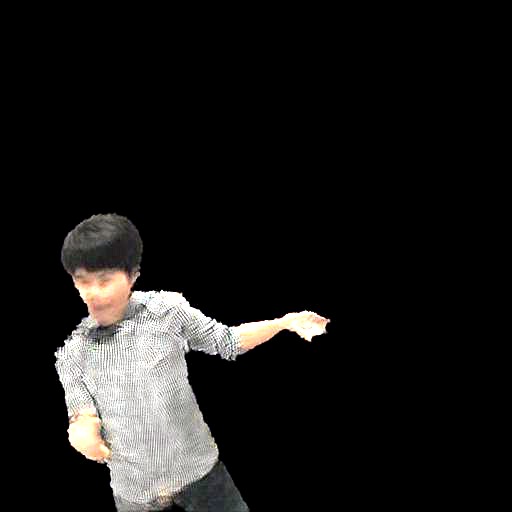}
      \caption{V2V}
      \end{subfigure}
      \begin{subfigure}{0.25\columnwidth}
      \includegraphics[trim={0cm 0cm 5cm 2.5cm},clip,width=\textwidth]{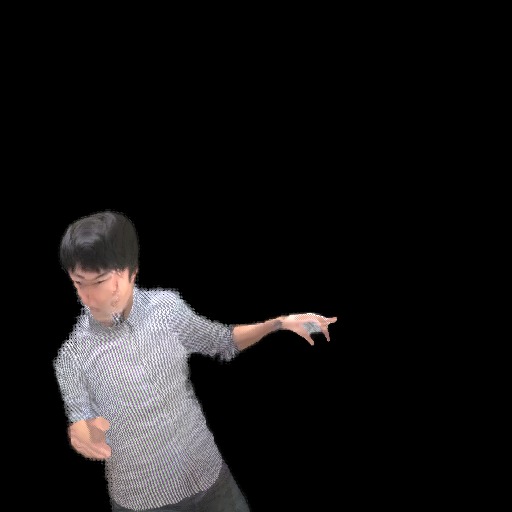}
      \caption{Proposed}
      \end{subfigure}

    \caption{Comparison of the rendering quality for the Direct, V2V and proposed methods on the \texttt{CMU1-6} and \texttt{CMU2-6} sequences. Images from six arbitrarily chosen cameras were used for training. We generate the views onto the hold-out cameras which were not used during training. The pose and camera in the lower right corner are in particular difficult for all the systems.}
    \label{fig:quads}
\end{figure*}

\paragraph{Baselines.} In the multi-video training scenario, we consider two other systems, against which ours is compared. First, we take the video-to-video (\textit{V2V})  system~\cite{Wang18}, using the authors' code with minimal modifications that lead to improved performance. We provide it with the same input as ours, and we use images with blacked-out background (according to our segmentation) as desired output. On the \texttt{CMU1-6} task, we have also evaluated a model with DensePose results computed on the target frame given as input (alongside keypoints). Despite much stronger (oracle-type) conditioning, the performance of this model in terms of considered metrics has not improved in comparison with V2V that uses only body joints as input.

The video-to-video system employs several adversarial losses and an architecture different from ours. Therefore we consider a more direct ablation (\textit{Direct}), which has the same network architecture that predicts RGB color and mask directly, rather than via body part assignments/coordinates. The Direct system is trained using the same losses and in the same protocol as ours. 

As for the single video case, two baseline systems, against which ours is compared, were considered. On our own captured sequences, we compare our system against video-to-video (\textit{V2V}) system ~\cite{Wang18}, whereas on sequences from \cite{Alldieck18a} we provide a qualitative comparison against the system of \cite{Alldieck18a}.   

\paragraph{Multi-video comparison.} We compare the three systems (\textit{ours}, \textit{V2V}, \textit{Direct}) in \texttt{CMU1-16}, \texttt{CMU2-16}, \texttt{CMU1-6}, \texttt{CMU2-6}. Using the hold-out sequences/motions, we then evaluated two popular metrics, namely  structured self-similarity (SSIM) and Frechet Inception Distance (FID) between the results of each system and the hold-out frames (with background removed using our segmentation algorithm). Our method outperforms the other two in terms of SSIM and underperforms V2V in terms of FID. Representative examples are shown in \fig{quads}.

We have also performed user study using a crowd-sourcing website, where the users were shown the results of ours and one of the other two systems on either side of the ground truth image and were asked to pick a better match to the middle image. In the side-by-side comparison, the results of our method were always preferred by the majority of crowd-sourcing users.
We note that our method suffers from a disadvantage both in the quantitative metrics and in the user comparison, since it averages out lighting from different viewpoints. The more detailed quantitative comparison is presented in \tab{tables}. 

\begin{figure*}[h!]
    \newlength{\wid}
    \setlength{\wid}{0.143\textwidth}
    \captionsetup[subfigure]{labelformat=empty}
    \centering
       \begin{subfigure}{\wid}
      \includegraphics[trim={5cm 0cm 5cm 17cm}, clip, width=\textwidth]{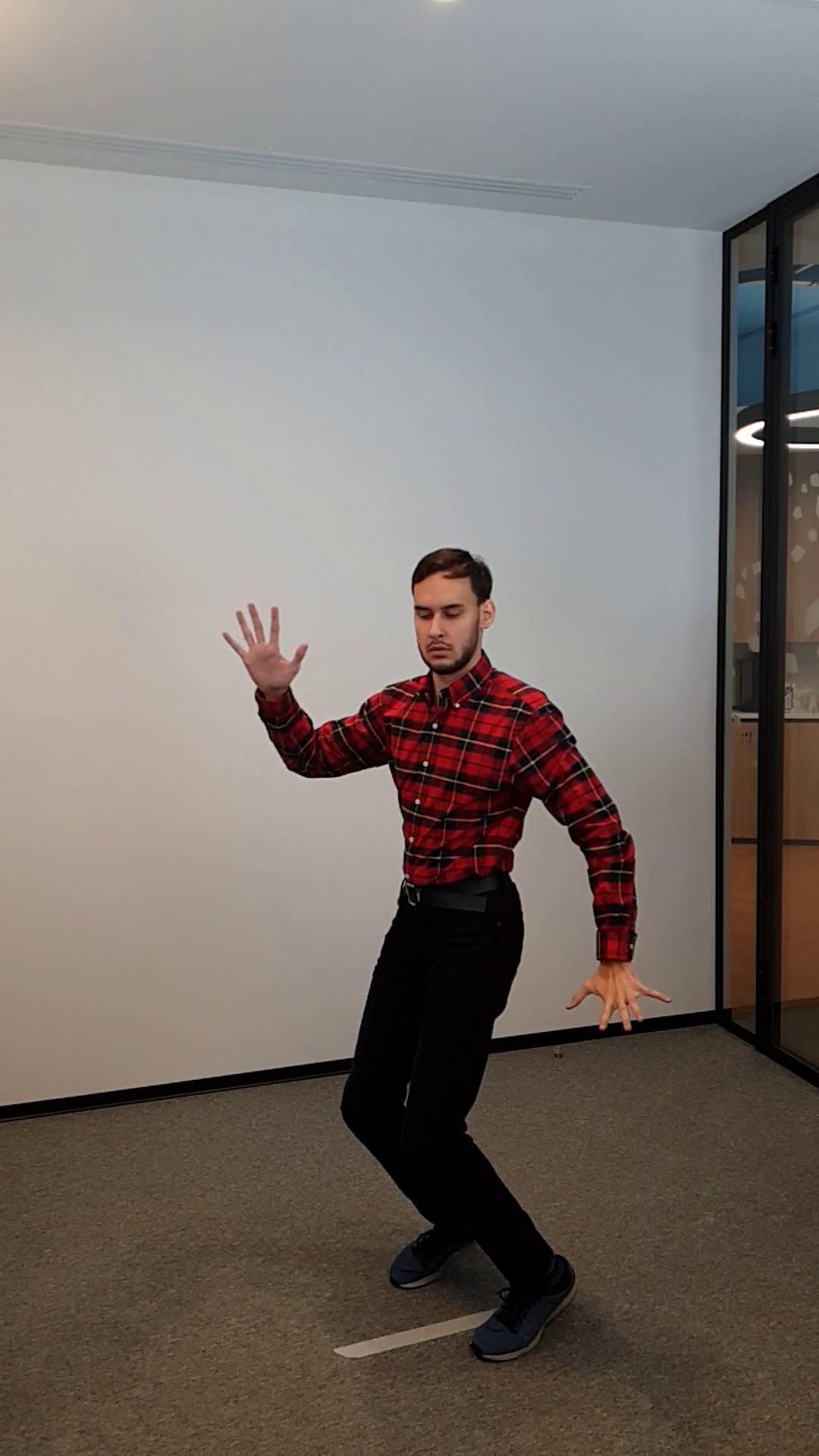} 
      \end{subfigure}
      \begin{subfigure}{\wid}
      \includegraphics[trim={5cm 0cm 5cm 17cm}, clip,width=\textwidth]{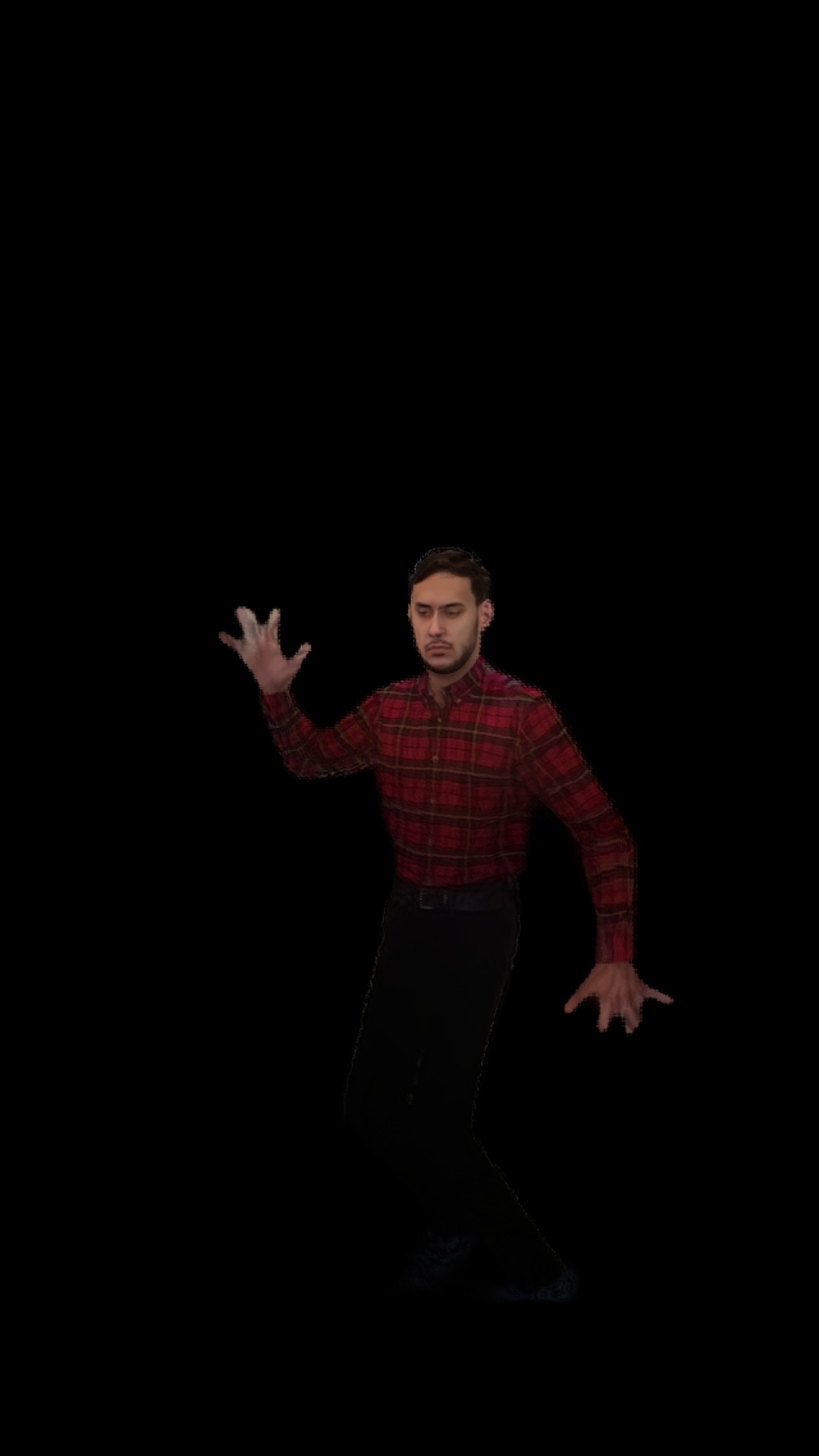} 
      \end{subfigure}
      \begin{subfigure}{\wid}
      \includegraphics[width=\textwidth,trim={5cm 0cm 5cm 17cm},clip]{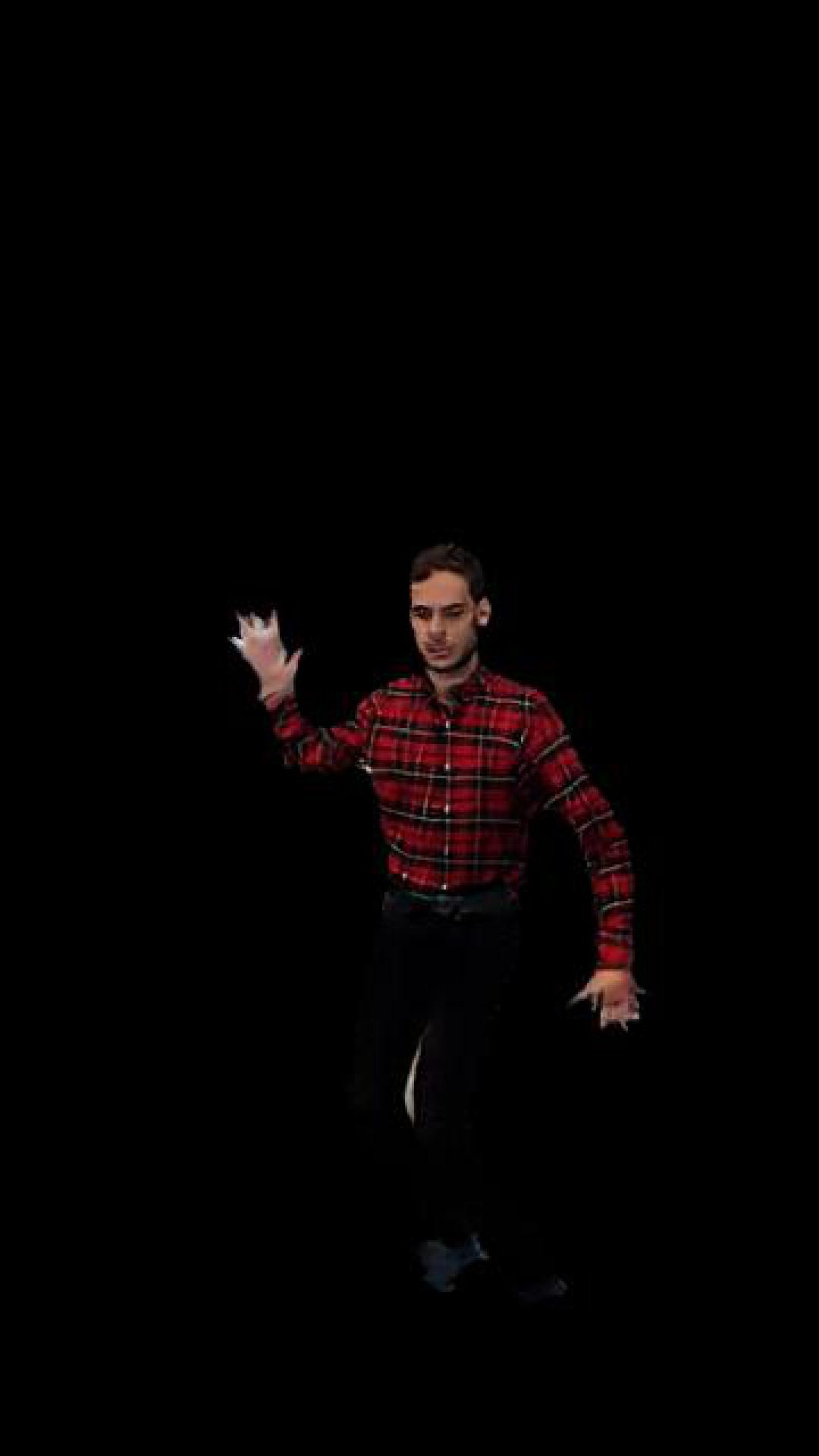}
      \end{subfigure}\hspace*{0.03em}
      \begin{subfigure}{\wid}
      \includegraphics[trim={8cm 0cm 2cm 17cm}, clip, width=\textwidth]{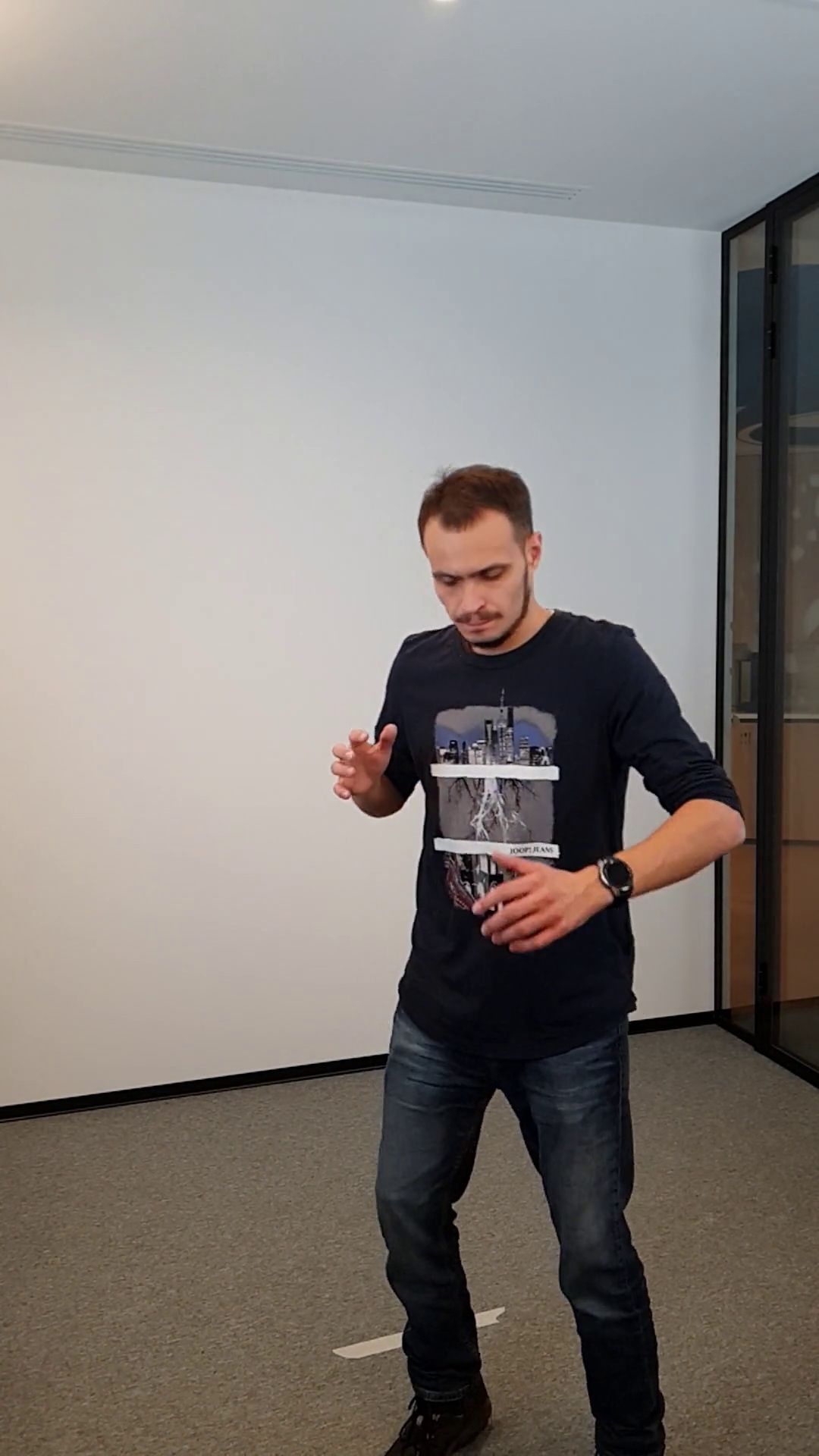} 
      \end{subfigure}
      \begin{subfigure}{\wid}
      \includegraphics[trim={8cm 0cm 2cm 17cm}, clip, width=\textwidth]{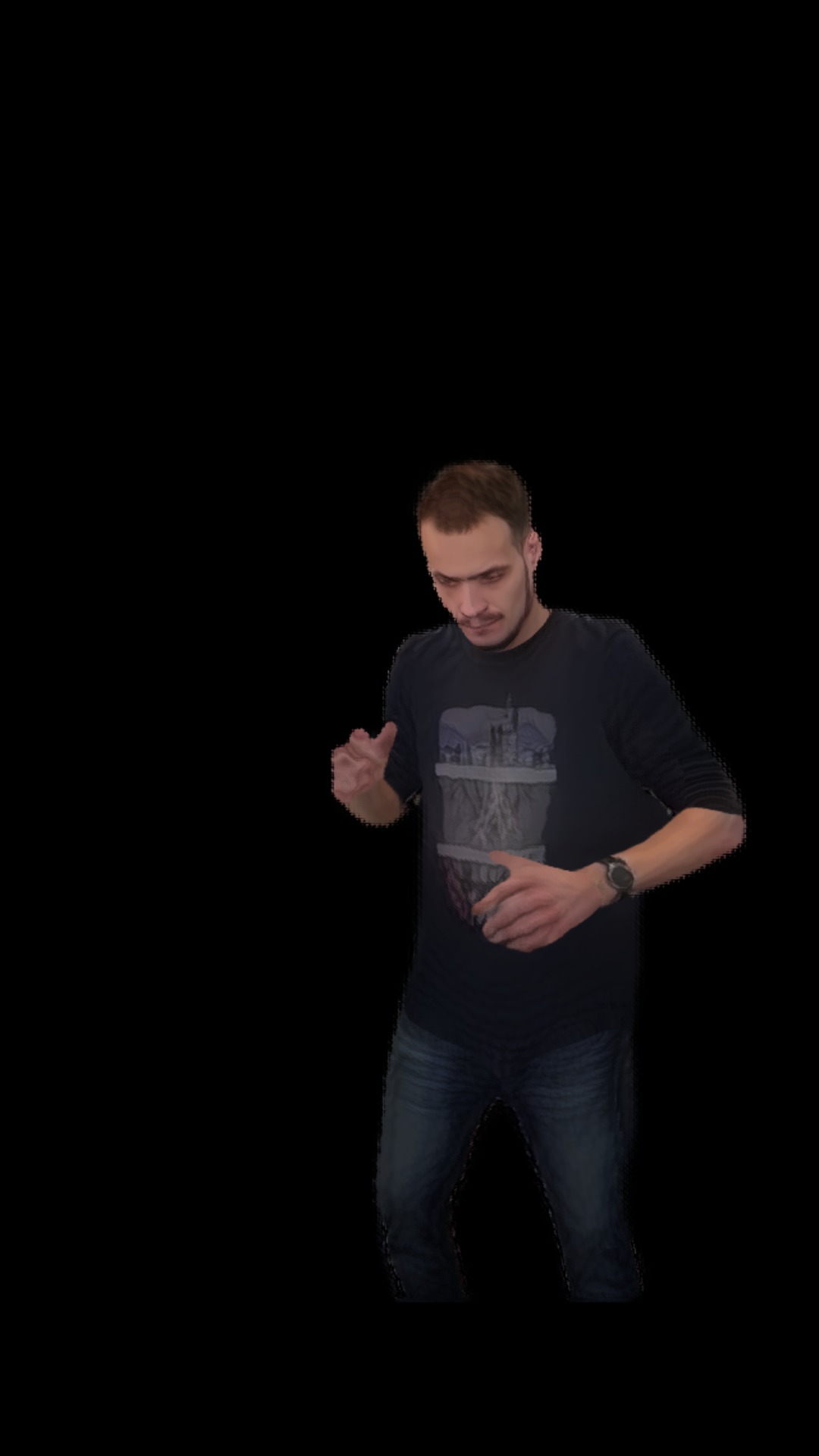} 
      \end{subfigure}
      \begin{subfigure}{\wid}
      \includegraphics[width=\textwidth,trim={8cm 0cm 2cm 17cm},clip]{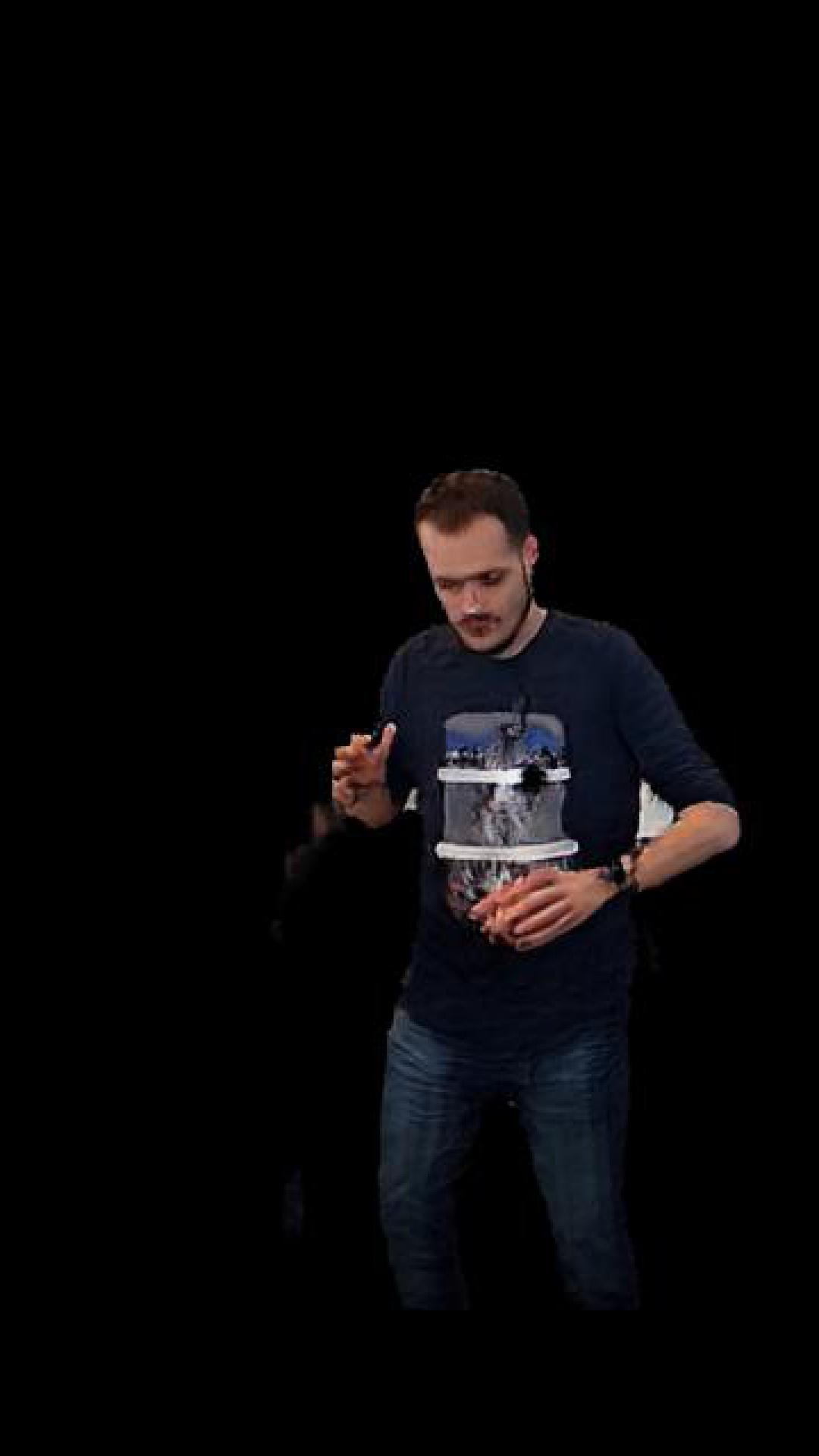}
      \end{subfigure}\\
      \begin{subfigure}{\wid}
      \includegraphics[trim={10cm 0cm 0cm 17cm}, clip,width=\textwidth]{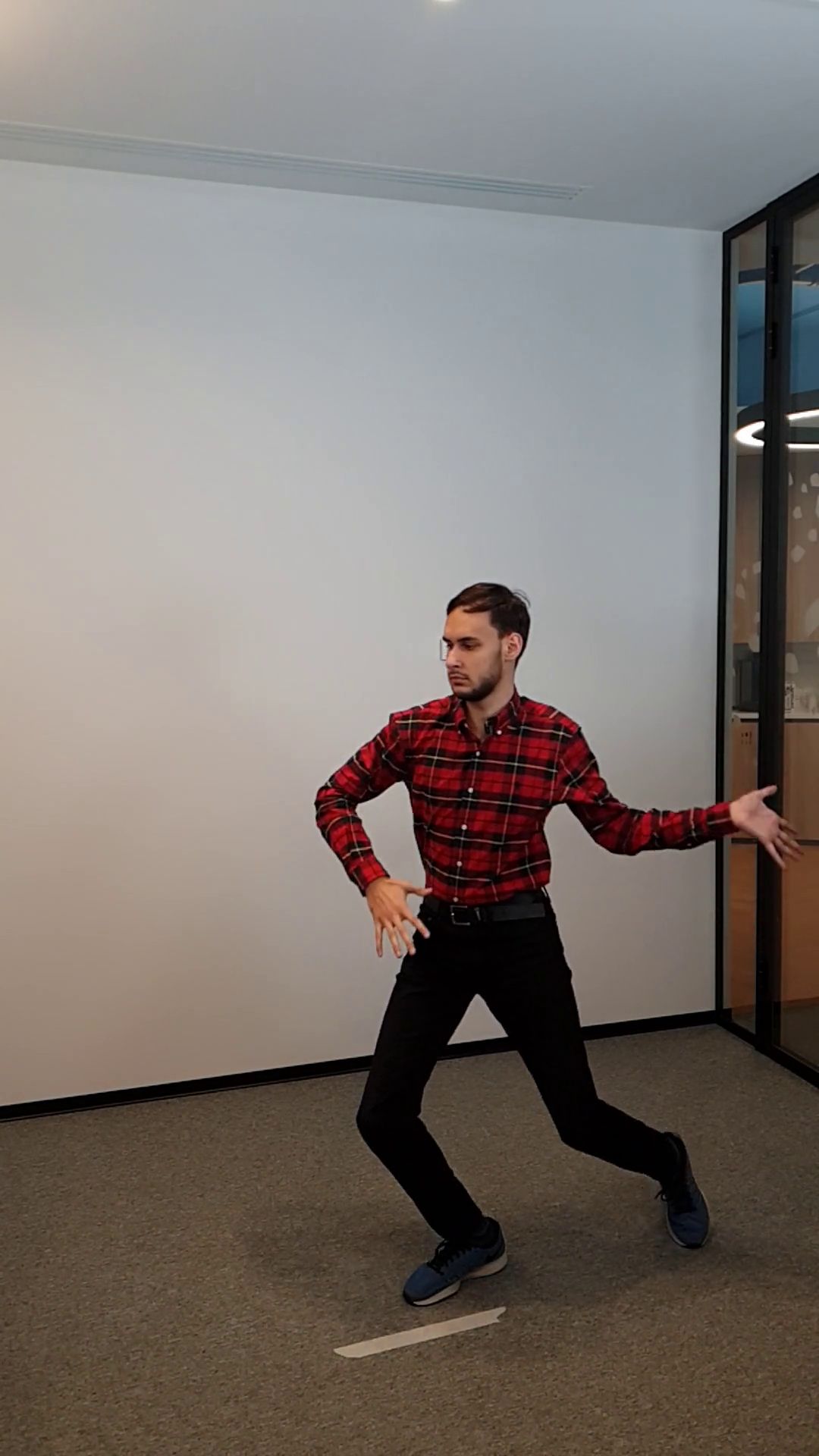} 
      \caption{GT}
      \end{subfigure}
      \begin{subfigure}{\wid}
      \includegraphics[trim={10cm 0cm 0cm 17cm}, clip,width=\textwidth]{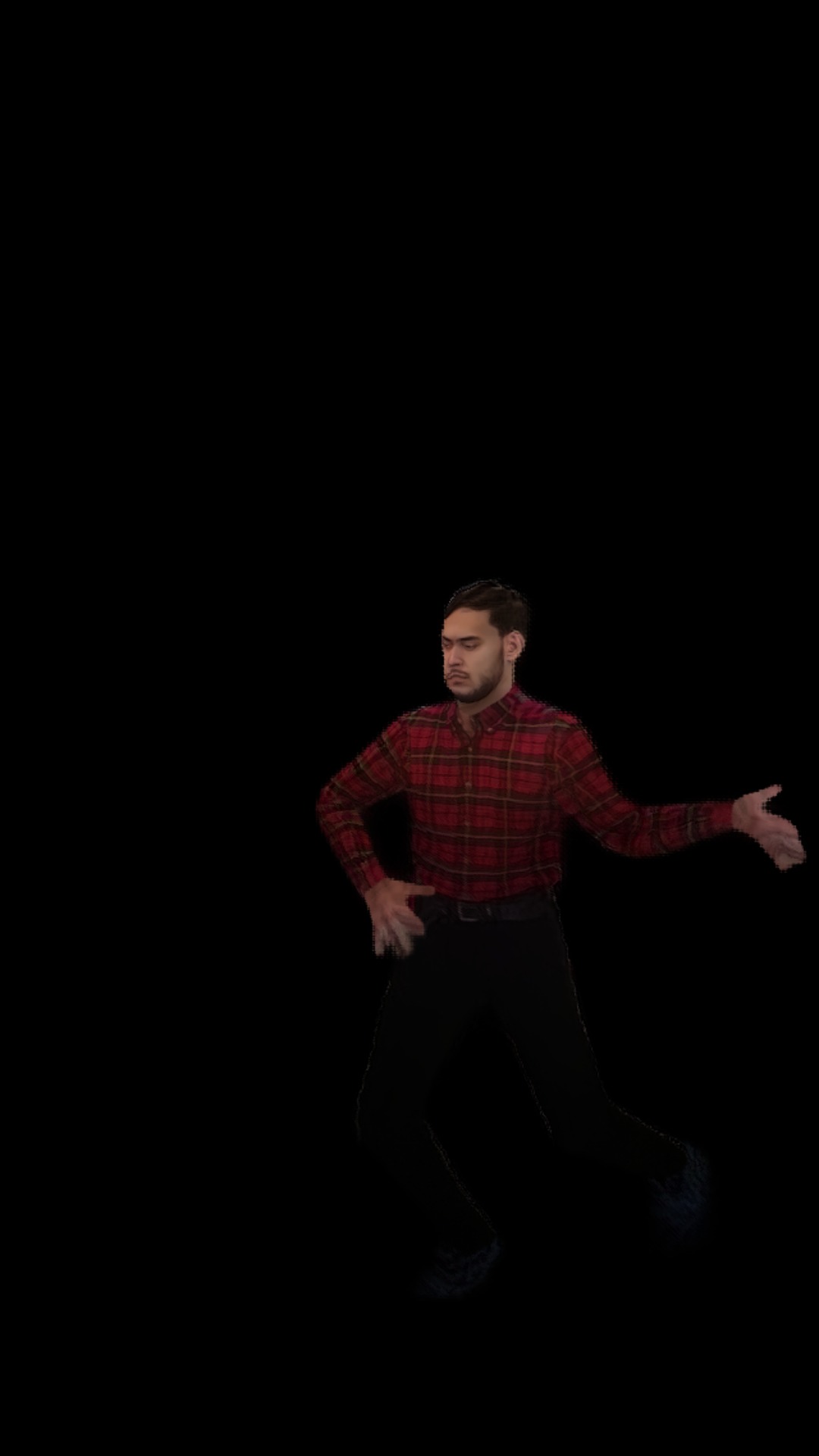} 
      \caption{Proposed}
      \end{subfigure}
      \begin{subfigure}{\wid}
      \includegraphics[width=\textwidth,trim={10cm 0cm 0cm 17cm},clip]{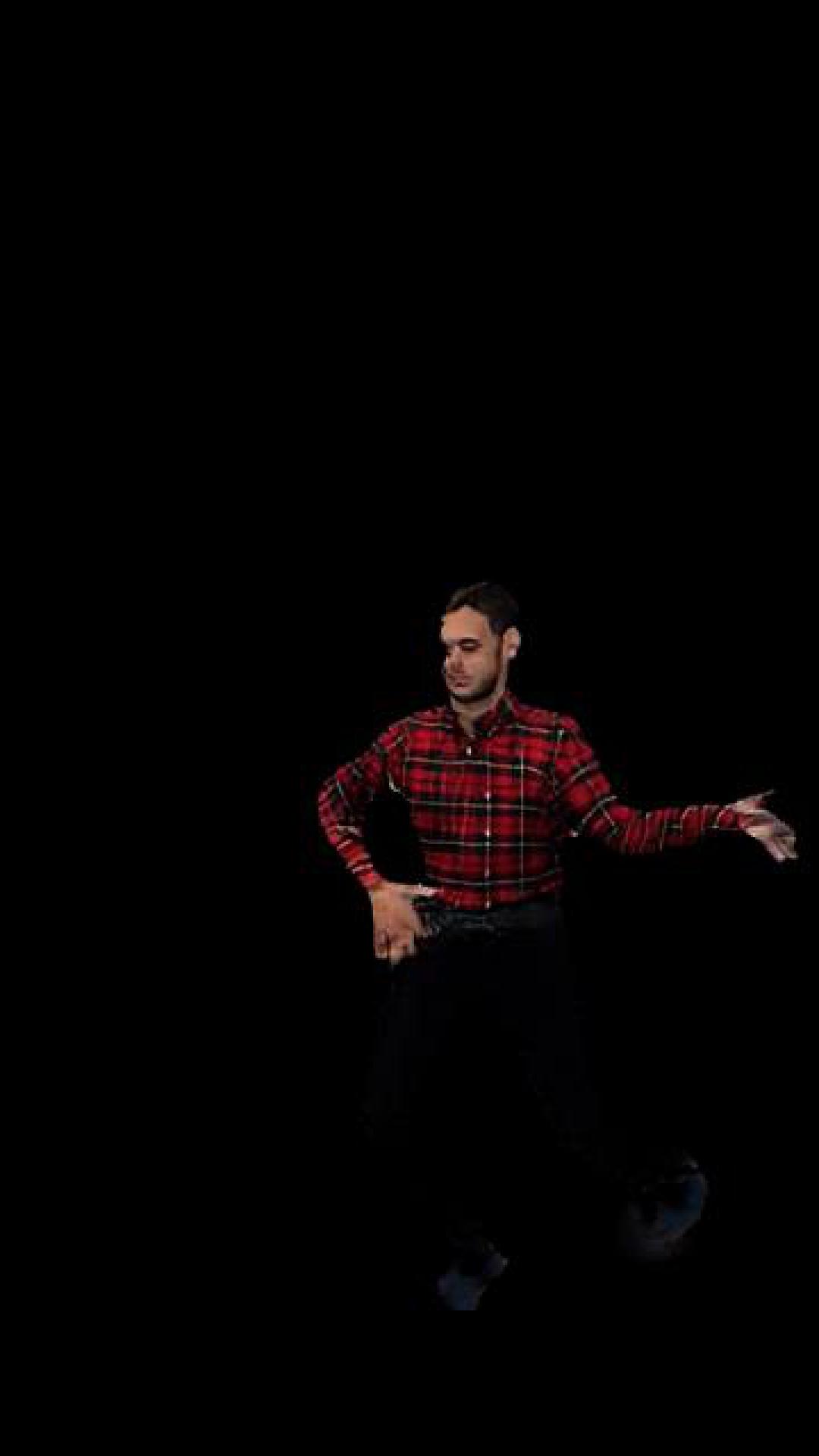} 
      \caption{V2V}
	  \end{subfigure}\hspace*{0.03em}
      \begin{subfigure}{\wid}
      \includegraphics[trim={5cm 0cm 5cm 17cm}, clip, width=\textwidth]{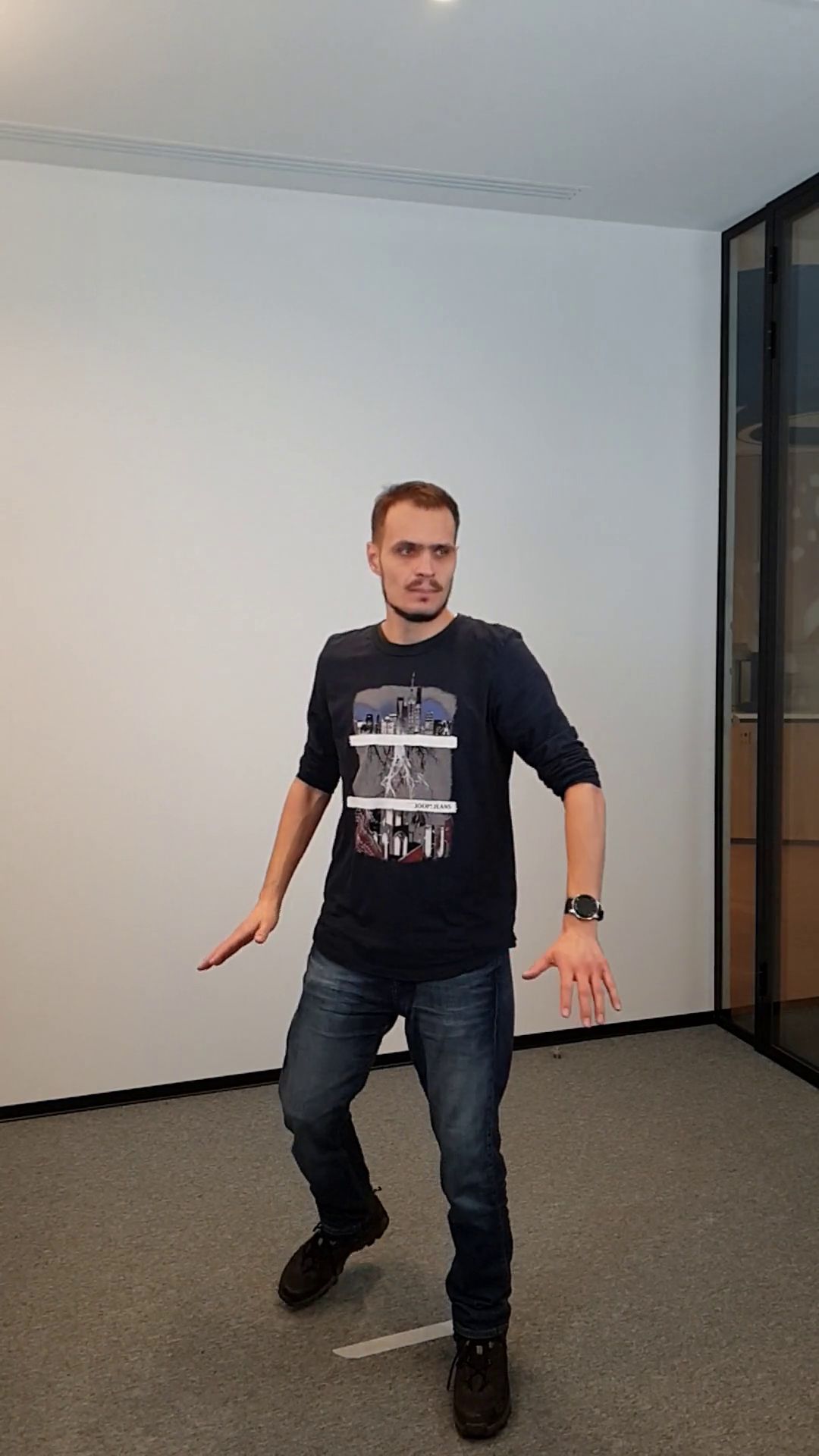} 
      \caption{GT}
      \end{subfigure}
      \begin{subfigure}{\wid}
      \includegraphics[trim={5cm 0cm 5cm 17cm}, clip, width=\textwidth]{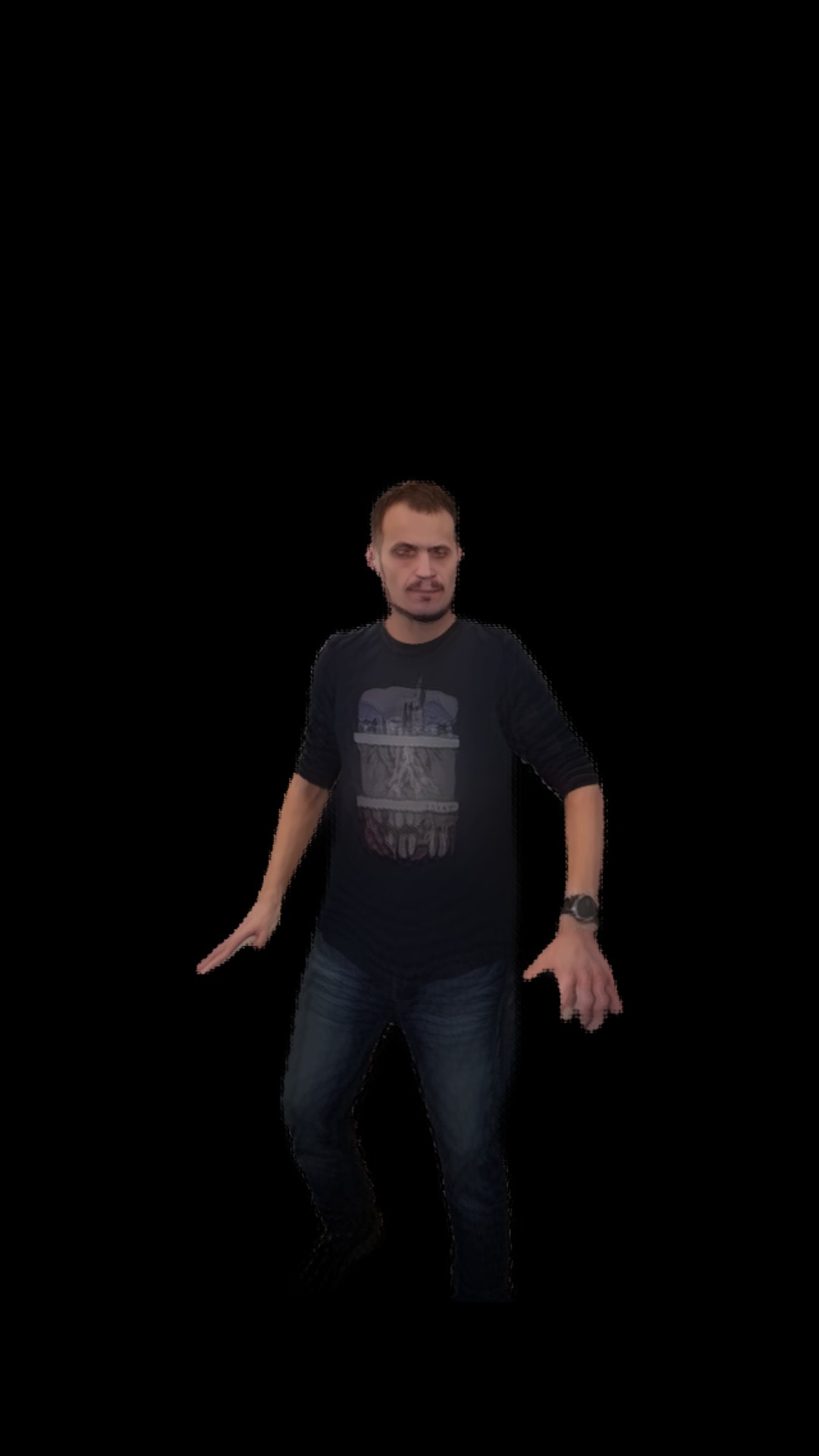} 
      \caption{Proposed}
      \end{subfigure}
      \begin{subfigure}{\wid}
      \includegraphics[width=\textwidth,trim={5cm 0cm 5cm 17cm},clip]{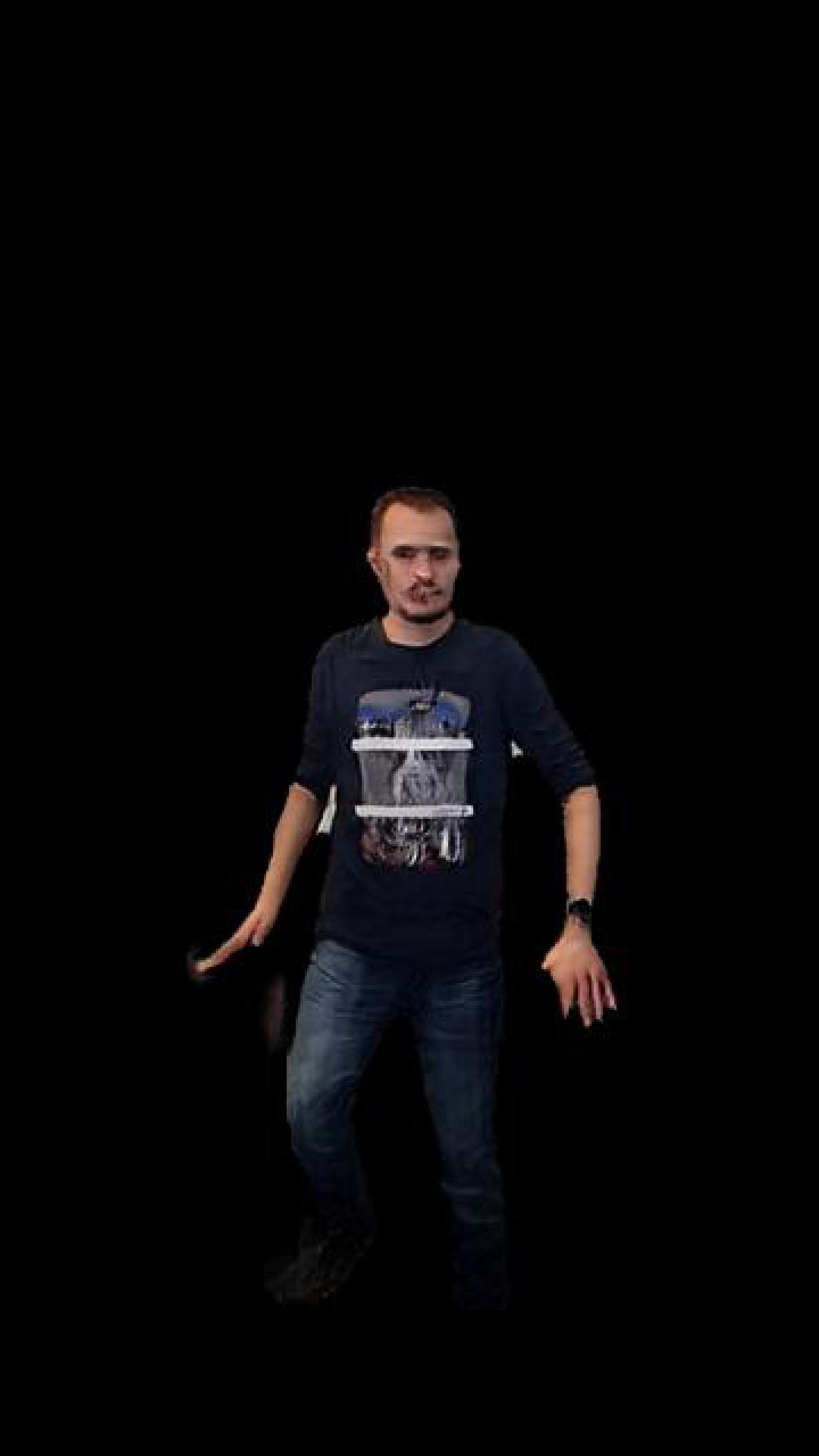} 
      \caption{V2V}
      \end{subfigure}
    \caption{Results comparison for our multi-view sequences using a hold-out camera. Textured Neural Avatars and the images produced by the video-to-video (V2V) system correspond to the same viewpoint. Both systems use a video from a single viewpoint for training. \textit{Electronic zoom-in recommended.}}
    \label{fig:triplets}
\end{figure*}

We show more qualitative examples of our method for a variety of models in \fig{assorti} and some qualitative comparisons with baselines in \fig{triplets}.

\paragraph{Single video comparisons.} We also evaluate our system in a single video case. We consider the scenario, where we train the model and transfer it to a new person by fitting it to a single video. We use single-camera videos from one of the cameras in our rig. We then evaluate the model (and V2V baseline) on a hold-out set of poses projected onto the camera from the other side of the rig (around $30^\circ$ away). We thus demonstrate that new models can be obtained using a single monocular video. For our models, we consider transferring from \texttt{CMU1-16}.

We thus pretrain V2V and our system on \texttt{CMU1-16} and use the obtained weights of $g_{\phi}$ as initialization for fine-tuning to the single video in our dataset. The texture maps are initialized from scratch as described above. Evaluating on hold-out camera and motion highlighted strong advantage of our method. In the user study on two subjects, the result of our method has been preferred to V2V in 55\% and 65\% of the cases. We further compare our method and the system of \cite{Alldieck18a} on the sequences from \cite{Alldieck18a}. The qualitative comparison is shown in \fig{extra}. In addition, we generate an avatar from a YouTube video. In this set of experiments, the avatars were obtained by fine-tuning from the same avatar (shown in \fig{triplets}--left). Except for the considerable artefacts on hand parts, our system has generated avatars that can generalize to new pose despite very short video input (300 frames in the case of \cite{Alldieck18a}).

\section{Summary and Discussion}

We have presented textured neural avatar approach to model the appearance of humans for new camera views and new body poses. Our system takes the middle path between the recent generation of methods that use ConvNets to map the pose to the image directly, and the traditional approach that uses geometric modeling of the surface and superimpose the personalized texture maps. This is achieved by learning a ConvNet that predicts texture coordinates of pixels in the new view jointly with the texture within the end-to-end learning process. We demonstrate that retaining an explicit shape and texture separation helps to achieve better generalization than direct mapping approaches.

Our method suffers from certain limitations. The generalization ability is still limited, as it does not generalize well when a person is rendered at a scale that is considerably different from the training set (which can be partially addressed by rescaling prior to rendering followed by cropping/padding postprocessing). Furthermore, textured avatars exhibit strong artefacts in the presence of pose estimation errors on hands and faces. Finally, our method assumes constancy of the surface color and ignores lighting effects. This can be potentially addressed by making our textures view- and lighting-dependent~\cite{Debevec98,Lombardi18}.


\FloatBarrier

{\small
\bibliographystyle{ieee_fullname}
\bibliography{refs}
}
\end{document}